\documentclass[conference]{IEEEtran}
\IEEEoverridecommandlockouts

\usepackage{cite}
\usepackage{amsmath,amssymb,amsfonts}
\usepackage{graphicx}
\usepackage{textcomp}
\usepackage[table]{xcolor}
\usepackage{booktabs}
\usepackage{multirow}
\usepackage{makecell}
\usepackage{xspace}
\usepackage{algorithm}
\usepackage{algpseudocode}
\usepackage{cuted}
\usepackage[hidelinks]{hyperref}

\algrenewcommand\algorithmicrequire{\textbf{Input:}}
\algrenewcommand\algorithmicensure{\textbf{Output:}}

\newcommand{\name}{Zeva\xspace}

\newcommand{\RoboCasaResultsTable}{%
\begin{table}[!t]
    \centering
    \caption{Results on the RoboCasa365-Atomic5 (Atomic5) benchmark~\cite{robocasa365}.
    We report task success rate (SR) over 50 randomized episodes and the
    macro-average. Best results are in \textbf{bold}.}
    \label{tab:exp_robocasa}
    \renewcommand{\arraystretch}{1.08}
    \setlength{\tabcolsep}{2.5pt}
    \scriptsize
    \resizebox{\linewidth}{!}{%
    \begin{tabular}{l|cccccc}
        \toprule[1.2pt]
        \textbf{Method}
        & \makecell{\textbf{TurnOn}\\\textbf{ElectricKettle}}
        & \makecell{\textbf{CloseToaster}\\\textbf{OvenDoor}}
        & \makecell{\textbf{TurnOn}\\\textbf{Microwave}}
        & \makecell{\textbf{CoffeeSetup}\\\textbf{Mug}}
        & \makecell{\textbf{OpenStand}\\\textbf{MixerHead}}
        & \textbf{Avg.} \\
        \midrule
        LingBot-VA~\cite{lingbot_va} & 63 & 60 & 70 & 35 & 48 & 55.2 \\
        Xiaomi-Robotics-1~\cite{xiaomi_robotics_1} & 17 & 16 & 82 & 6 & 35 & 31.2 \\
        $\tau_0$-WM~\cite{tau0_wm} & 15 & 32 & 63 & 12 & 23 & 29.0 \\
        Fast-WAM~\cite{fast_wam} & \textbf{88} & 66 & 60 & \textbf{54} & \textbf{94} & 72.4 \\
        Cosmos3-Nano~\cite{cosmos3} & 76 & 74 & 76 & 6 & 92 & 64.8 \\
        \midrule
        \rowcolor[HTML]{EAE6ED}
        \textbf{Zeva} & 78 & \textbf{86} & \textbf{84} & 44 & 92 & \textbf{76.8} \\
        \bottomrule[1.2pt]
    \end{tabular}}
\end{table}}

\newcommand{\RealWorldResultsTable}{%
\begin{table*}[!t]
    \centering
    \caption{Real-world evaluation on ChemLab-Evo. The left block reports
    success rate (SR, \%) over 20 randomized episodes, with macro-averages for
    each difficulty level. The right block reports the average percentage of
    ordered task stages completed on the two long-horizon tasks. Best process
    scores are in \textbf{bold}.}
    \label{tab:exp_real}
    \label{tab:long-horizon-score}
    \renewcommand{\arraystretch}{1.10}
    \setlength{\tabcolsep}{2.0pt}
    \scriptsize
    \resizebox{\textwidth}{!}{%
    \begin{tabular}{@{}l|cccc|ccc|ccc|ccc@{}}
        \toprule[1.2pt]
        \multirow{3}{*}{\textbf{Method}}
        & \multicolumn{10}{c|}{\textbf{Success Rate (SR, \%)}}
        & \multicolumn{3}{c}{\textbf{Process Score (\%)}} \\
        \cmidrule(lr){2-11}\cmidrule(lr){12-14}
        & \multicolumn{4}{c|}{\textbf{Level 1 (Atomic)}}
        & \multicolumn{3}{c|}{\textbf{Level 2 (Short-sequence)}}
        & \multicolumn{3}{c|}{\textbf{Level 3 (Complex)}}
        & \multicolumn{3}{c}{\textbf{Long-horizon}} \\
        \cmidrule(lr){2-5}\cmidrule(lr){6-8}\cmidrule(lr){9-11}\cmidrule(lr){12-14}
        & \makecell{\textbf{Pick Up}\\\textbf{Test Tube}}
        & \makecell{\textbf{Place}\\\textbf{Beaker}}
        & \makecell{\textbf{Pour}\\\textbf{Water}}
        & \makecell{\textbf{Avg.}\\\textbf{(\%)}}
        & \textbf{Titration}
        & \makecell{\textbf{Prepare Salt}\\\textbf{Solution}}
        & \makecell{\textbf{Avg.}\\\textbf{(\%)}}
        & \makecell{\textbf{Balance}\\\textbf{Weighing}}
        & \textbf{Extraction}
        & \textbf{Avg.}
        & \makecell{\textbf{Balance}\\\textbf{Weighing}}
        & \textbf{Extraction}
        & \textbf{Avg.} \\
        \midrule
        LingBot-VA~\cite{lingbot_va}
        & 80 & 70 & 65 & 71.7
        & 55 & 75 & 65.0
        & 5 & 0 & 2.5
        & 28.75 & 40.00 & 34.38 \\
        Fast-WAM~\cite{fast_wam}
        & 85 & 70 & 75 & 76.7
        & 45 & 55 & 50.0
        & 0 & 0 & 0.0
        & 22.50 & 58.57 & 40.54 \\
        $\pi_{0.5}$~\cite{pi05}
        & 95 & 60 & 65 & 73.3
        & 60 & 50 & 55.0
        & 0 & 0 & 0.0
        & 33.75 & 31.43 & 32.59 \\
        Cosmos3-Edge~\cite{cosmos3}
        & 80 & 70 & 60 & 70.0
        & 45 & 60 & 52.5
        & 0 & 0 & 0.0
        & 38.75 & 50.71 & 44.73 \\
        \midrule
        \rowcolor[HTML]{EAE6ED}
        \textbf{Zeva}
        & 100 & 70 & 80 & \textbf{83.3}
        & 70 & 70 & \textbf{70.0}
        & 5 & 10 & \textbf{7.5}
        & \textbf{47.50} & \textbf{67.14} & \textbf{57.32} \\
        \bottomrule[1.2pt]
    \end{tabular}}
\end{table*}}

\newcommand{\AblationResultsTable}{%
\begin{table}[H]
    \centering
    \caption{Real-world ablation of BIT and PIM on five ChemLab-Evo tasks.
    Checkmarks indicate enabled components. Entries report task-level SR (\%)
    over 20 randomized episodes; best results are in \textbf{bold}.}
    \label{tab:ablation}
    \renewcommand{\arraystretch}{1.05}
    \setlength{\tabcolsep}{1.8pt}
    \scriptsize
    \begin{tabular}{@{}cc|ccccc@{}}
        \toprule[1.2pt]
        \multicolumn{2}{c|}{\textbf{Method}}
        & \multirow{2}{*}{\makecell{\textbf{Pick Up}\\\textbf{Test Tube}}}
        & \multirow{2}{*}{\makecell{\textbf{Place}\\\textbf{Beaker}}}
        & \multirow{2}{*}{\makecell{\textbf{Pour}\\\textbf{Water}}}
        & \multirow{2}{*}{\textbf{Titr.}}
        & \multirow{2}{*}{\makecell{\textbf{Salt}\\\textbf{Solution}}} \\
        \cmidrule(lr){1-2}
        \textbf{BIT} & \textbf{PIM} & & & & & \\
        \midrule
        \rowcolor[HTML]{EAE6ED}
        $\checkmark$ & $\checkmark$ & \textbf{100} & \textbf{70}
        & \textbf{80} & \textbf{70} & \textbf{70} \\
        & $\checkmark$ & 80 & 50 & 50 & 40 & 55 \\
        $\checkmark$ & & 85 & 60 & 70 & 60 & 50 \\
        & & 65 & 40 & 45 & 25 & 35 \\
        \bottomrule[1.2pt]
    \end{tabular}
\end{table}}

\newcommand{\CTERepresentationFigure}{%
\begin{figure}[H]
    \centering
    \setlength{\tabcolsep}{1.2pt}
    \begin{tabular}{@{}cc@{}}
        \includegraphics[width=0.49\columnwidth,trim=0 145 0 0,clip]
            {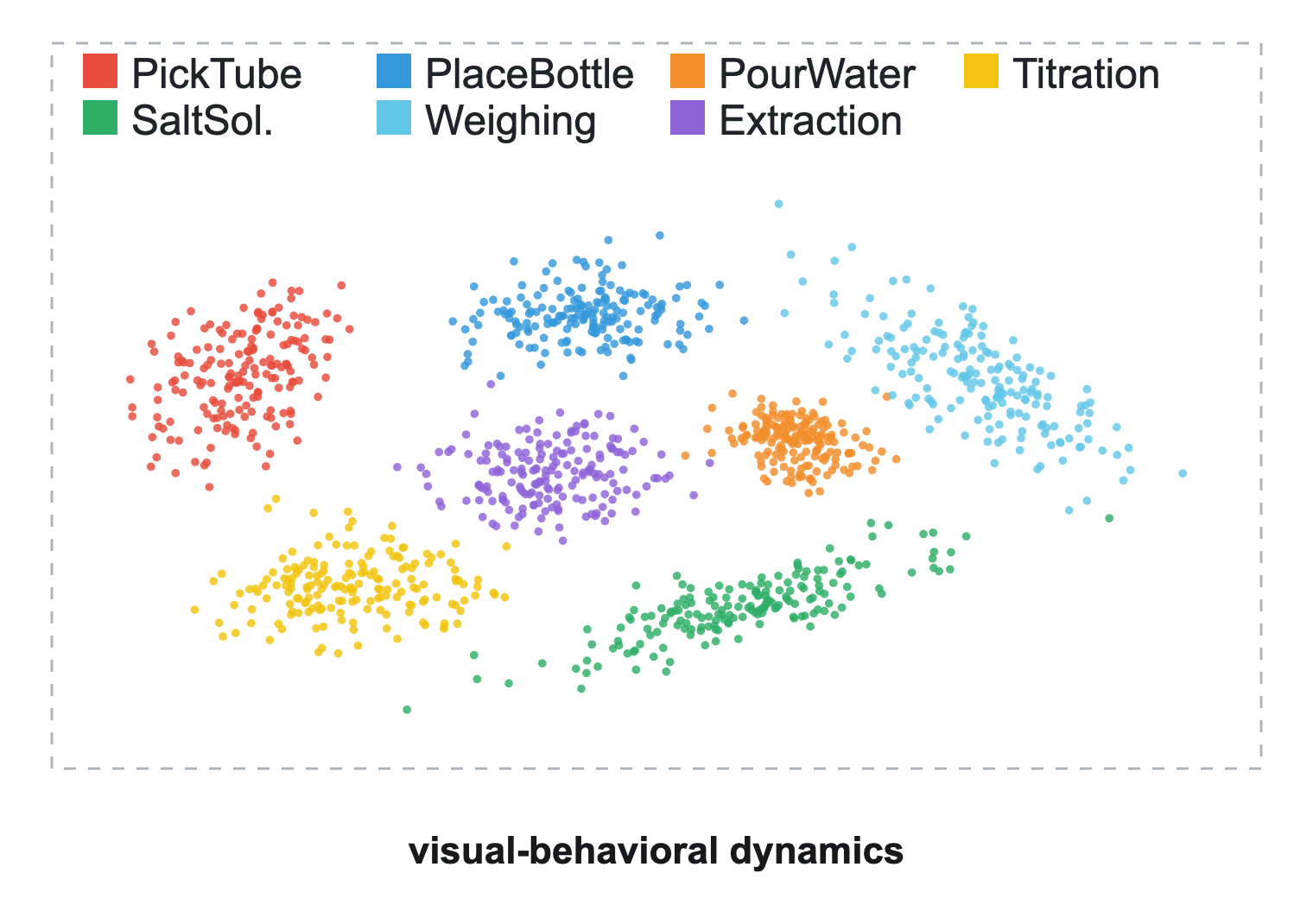} &
        \includegraphics[width=0.49\columnwidth,trim=0 145 0 0,clip]
            {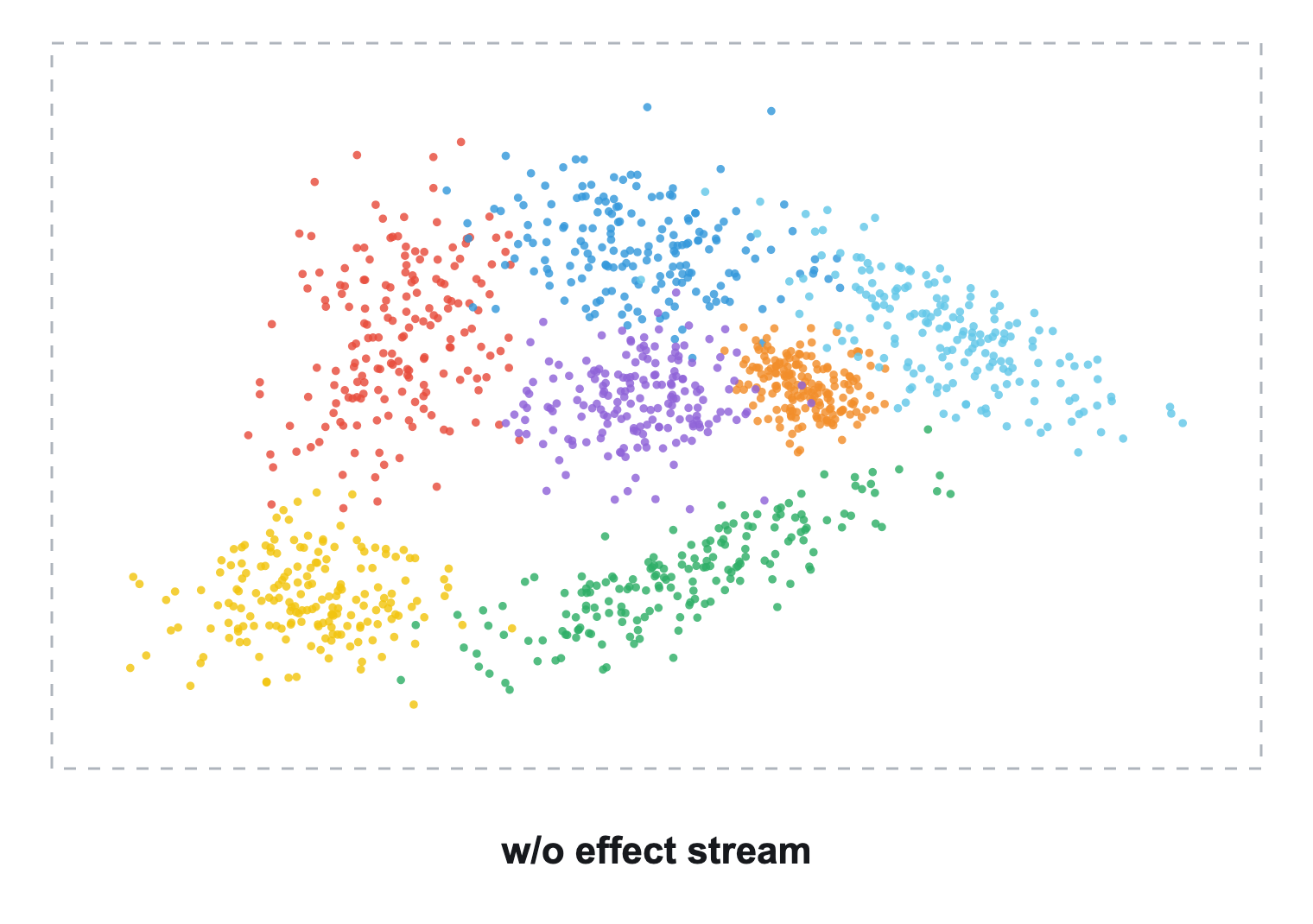} \\[-0.8mm]
        {\footnotesize\textbf{(a) Full CTE}} &
        {\footnotesize\textbf{(b) w/o Effect Stream}}
    \end{tabular}
    \caption{t-SNE visualization of CTE interaction embeddings across seven
    ChemLab-Evo tasks. Full CTE forms compact clusters, whereas removing the
    effect stream increases dispersion and cross-task overlap.}
    \label{fig:cte-tsne}
\end{figure}}

\newcommand{\failedattempt}[1]{%
    \fcolorbox{red!75!black}{white}{\includegraphics[width=0.22\columnwidth]{#1}}}
\newcommand{\successfulattempt}[1]{%
    \fcolorbox{green!55!black}{white}{\includegraphics[width=0.22\columnwidth]{#1}}}
\newcommand{\failedwideattempt}[1]{%
    \fcolorbox{red!75!black}{white}{\includegraphics[width=0.22\columnwidth,trim=228 0 228 0,clip]{#1}}}
\newcommand{\successfulwideattempt}[1]{%
    \fcolorbox{green!55!black}{white}{\includegraphics[width=0.22\columnwidth,trim=228 0 228 0,clip]{#1}}}

\newcommand{\CaseEvolutionFigure}{%
\begin{figure}[H]
    \centering
    \setlength{\tabcolsep}{0.5pt}
    \setlength{\fboxsep}{0pt}
    \setlength{\fboxrule}{1.1pt}
    \begin{tabular}{@{}>{\centering\arraybackslash}m{0.055\columnwidth}cccc@{}}
        & \multicolumn{4}{c}{\makebox[0.88\columnwidth]{\scriptsize
        \textcolor{gray!70!black}{Earlier\hfill$\longrightarrow$\hfill Later}}} \\[-0.3mm]
        {\scriptsize\textbf{(a)}} &
        \failedattempt{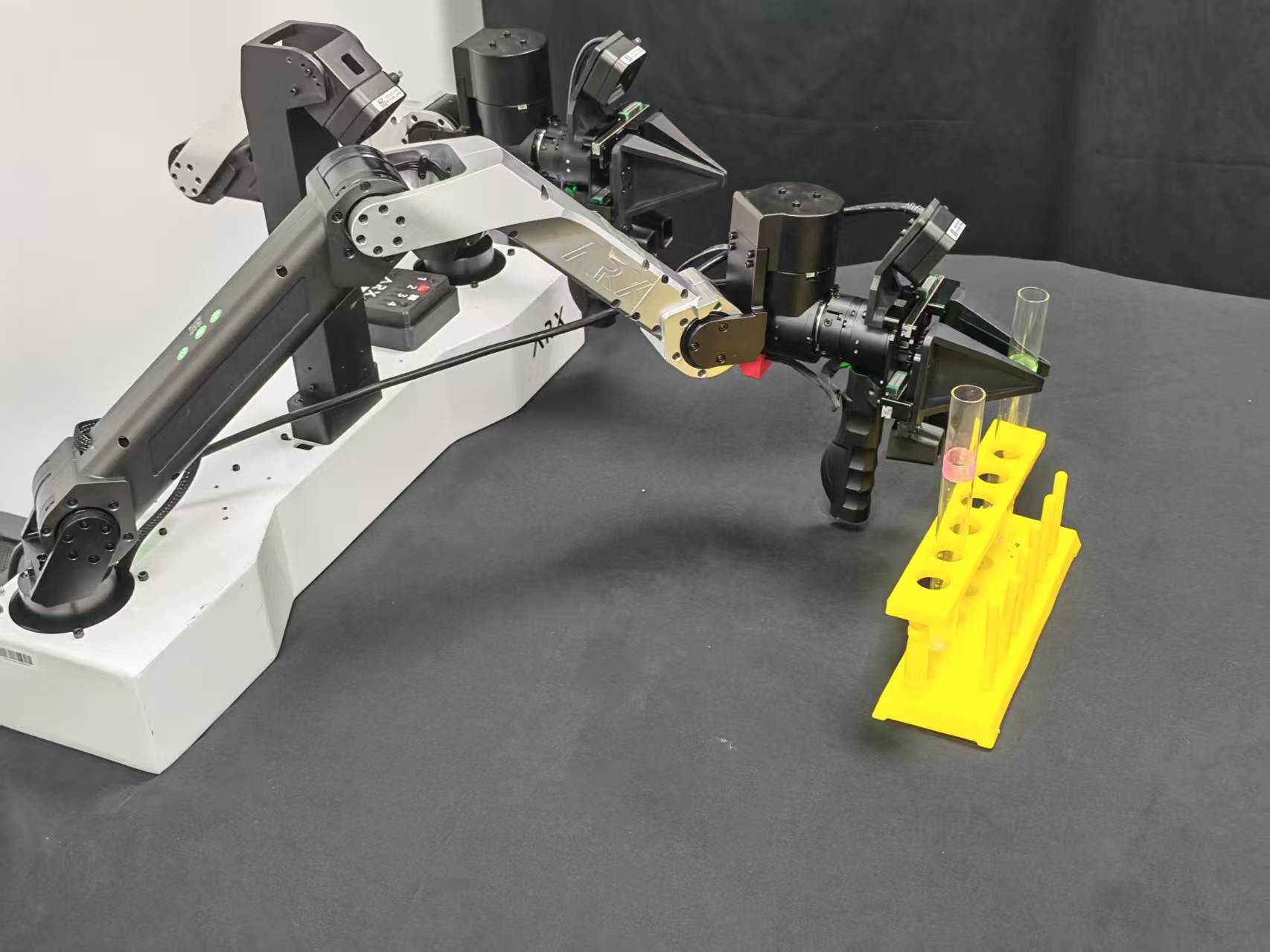} &
        \failedattempt{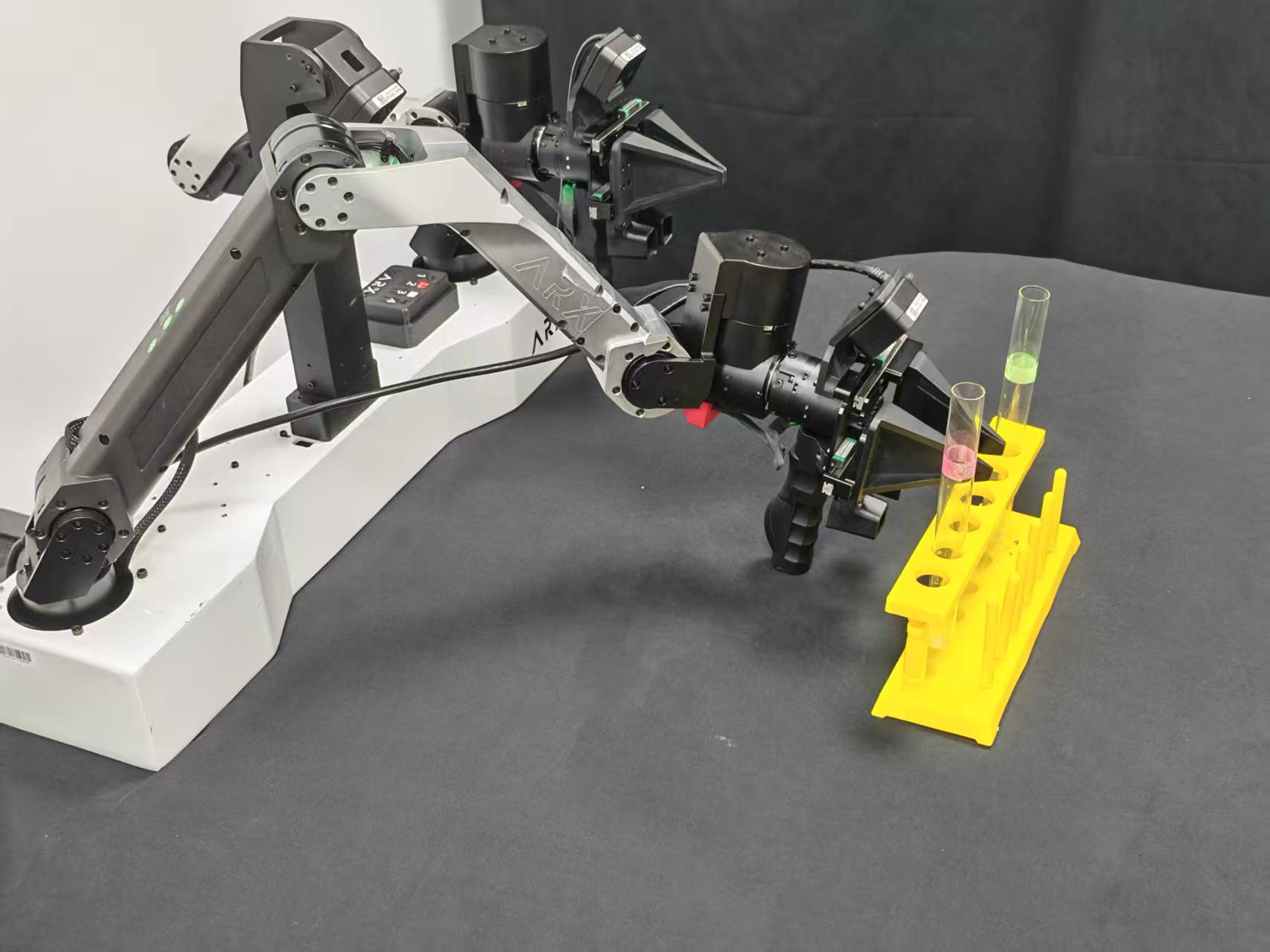} &
        \successfulattempt{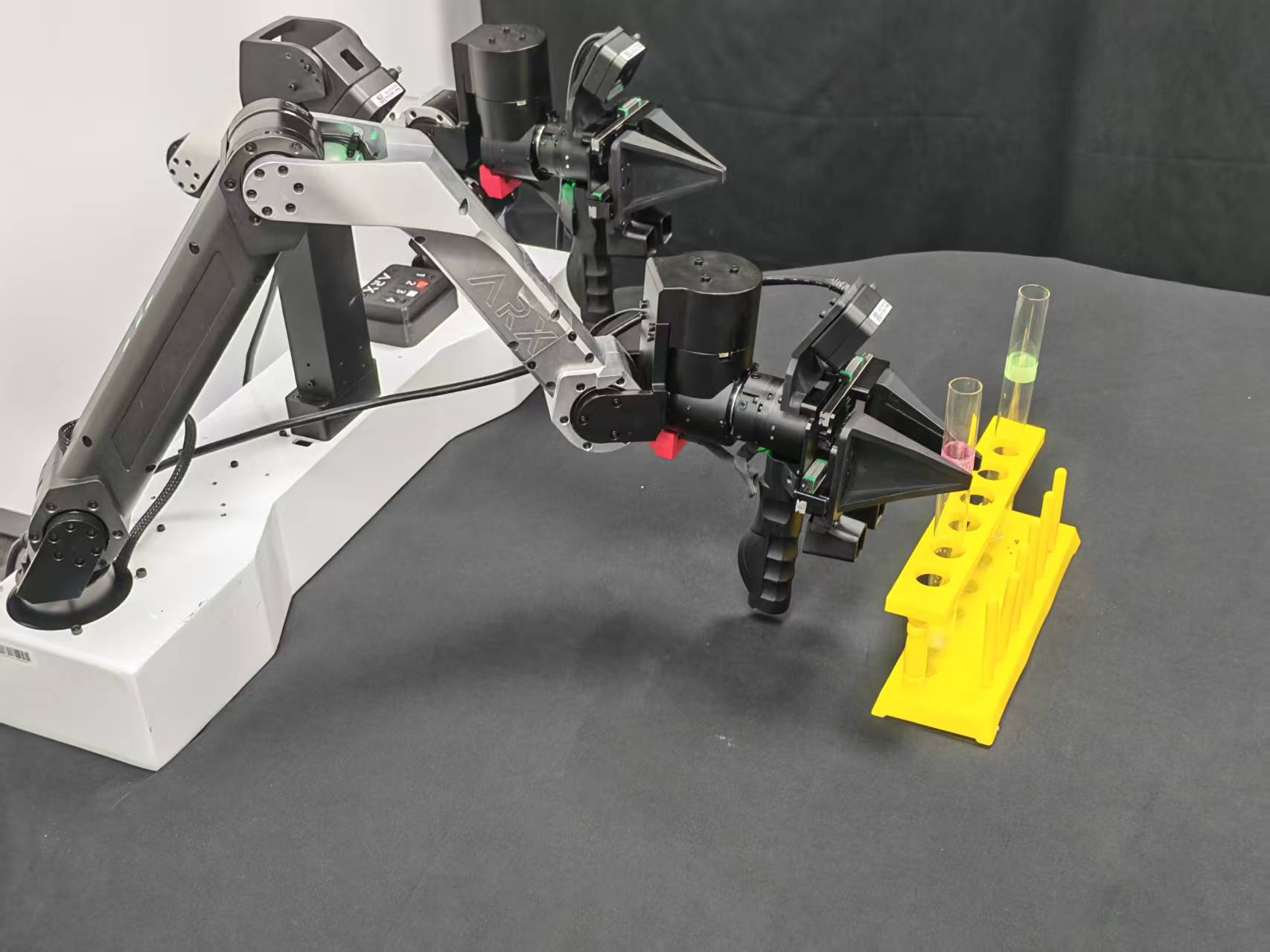} &
        \successfulattempt{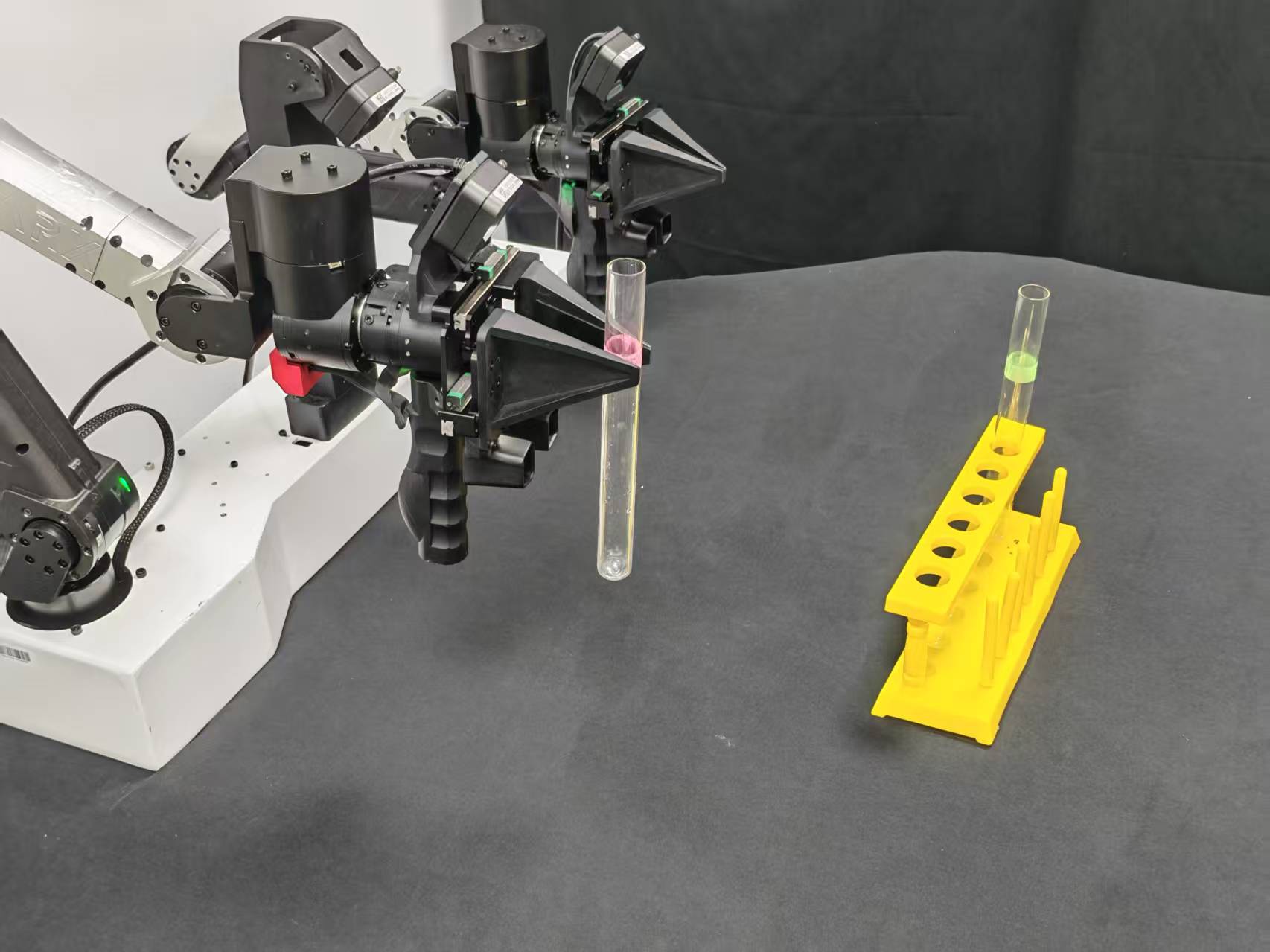} \\[0.8mm]
        {\scriptsize\textbf{(b)}} &
        \failedattempt{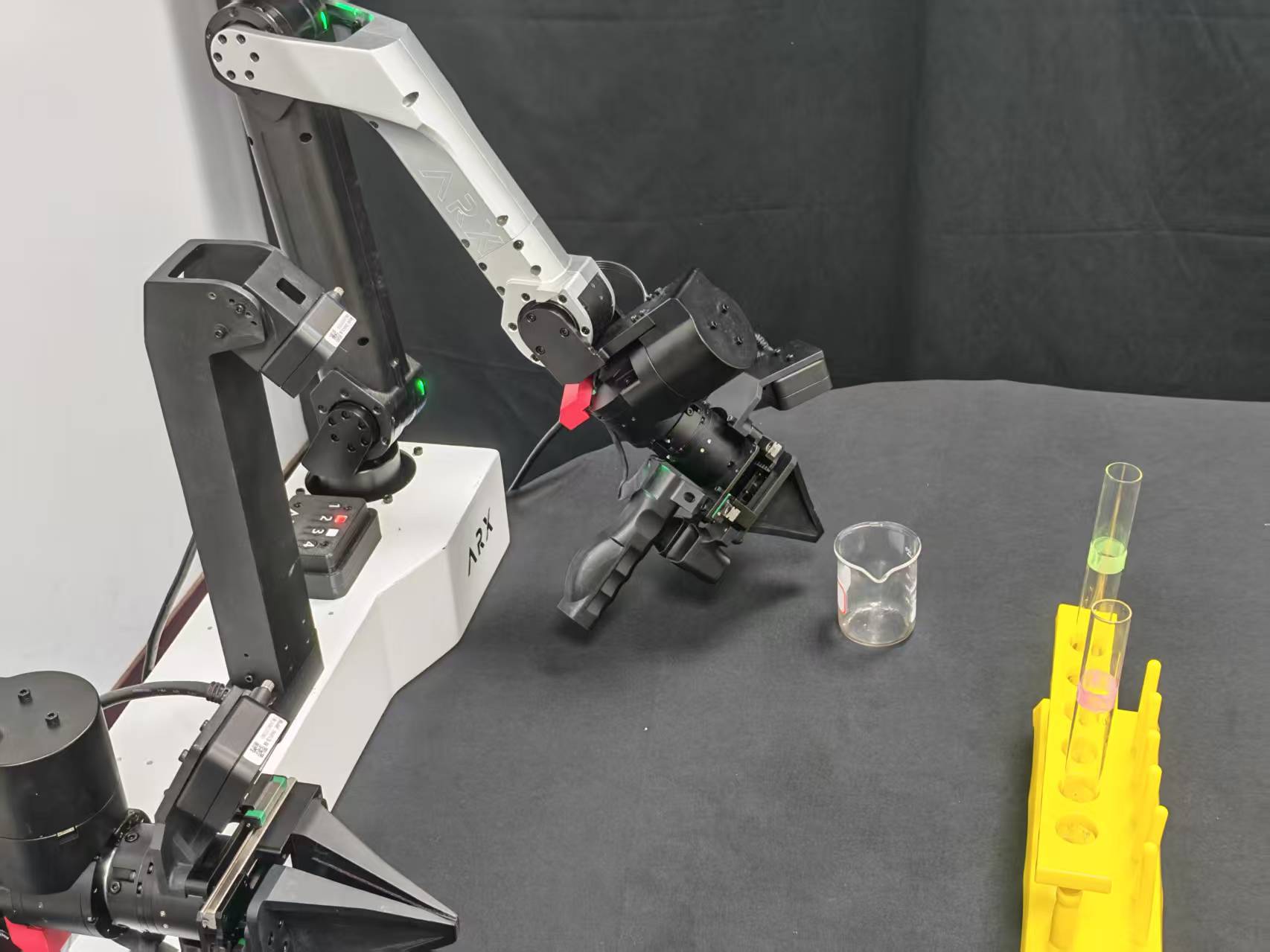} &
        \failedattempt{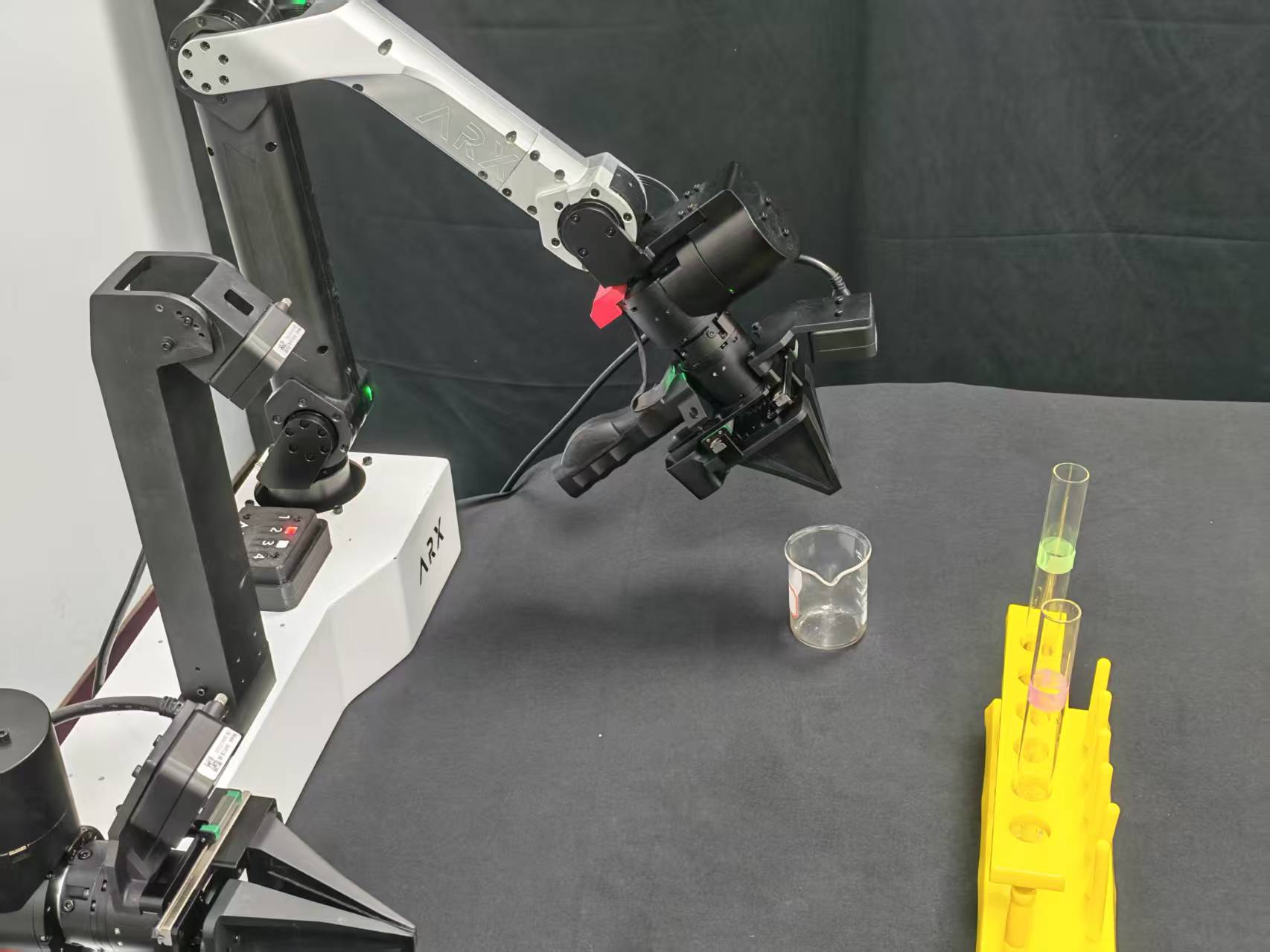} &
        \successfulattempt{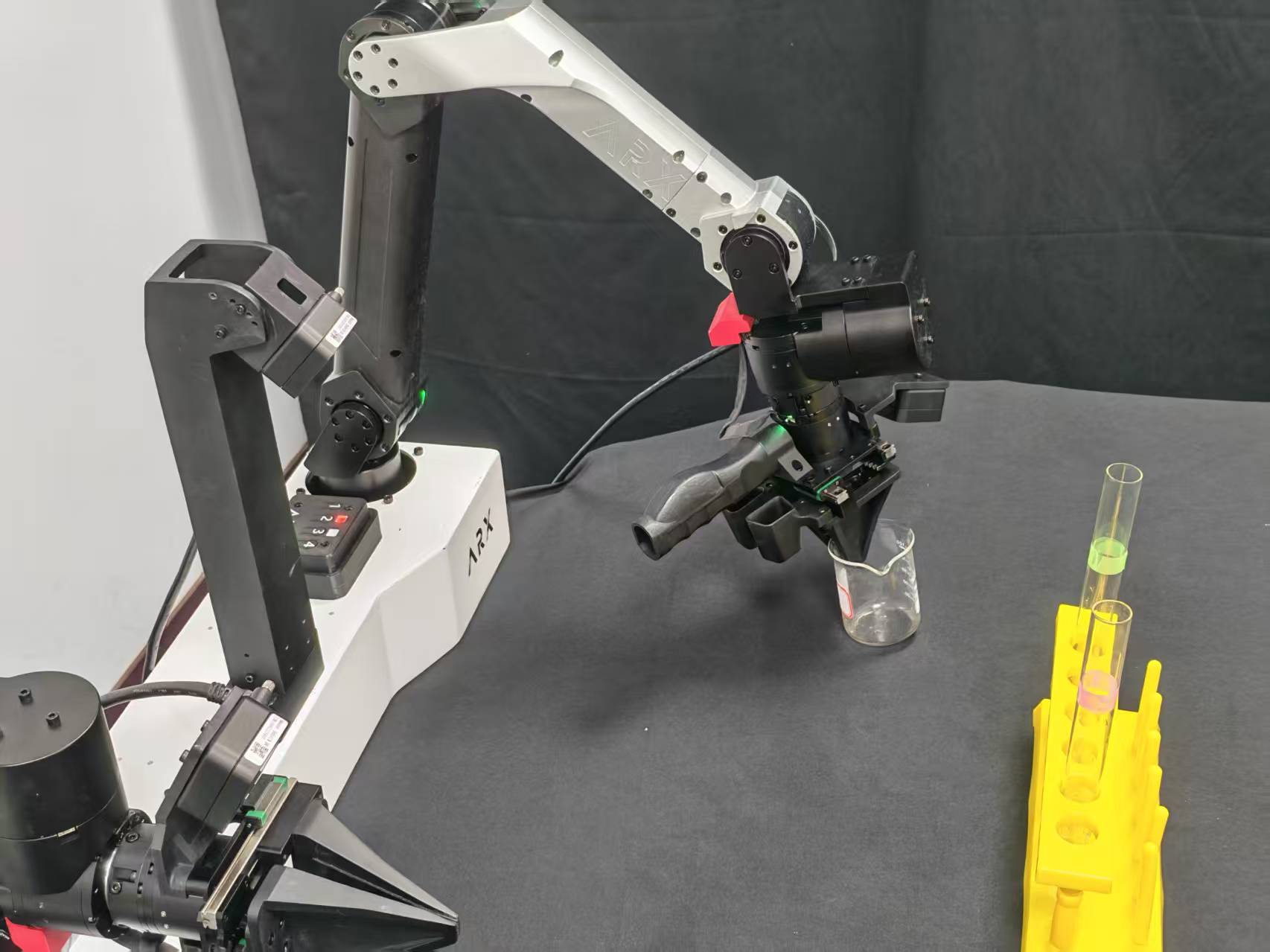} &
        \successfulattempt{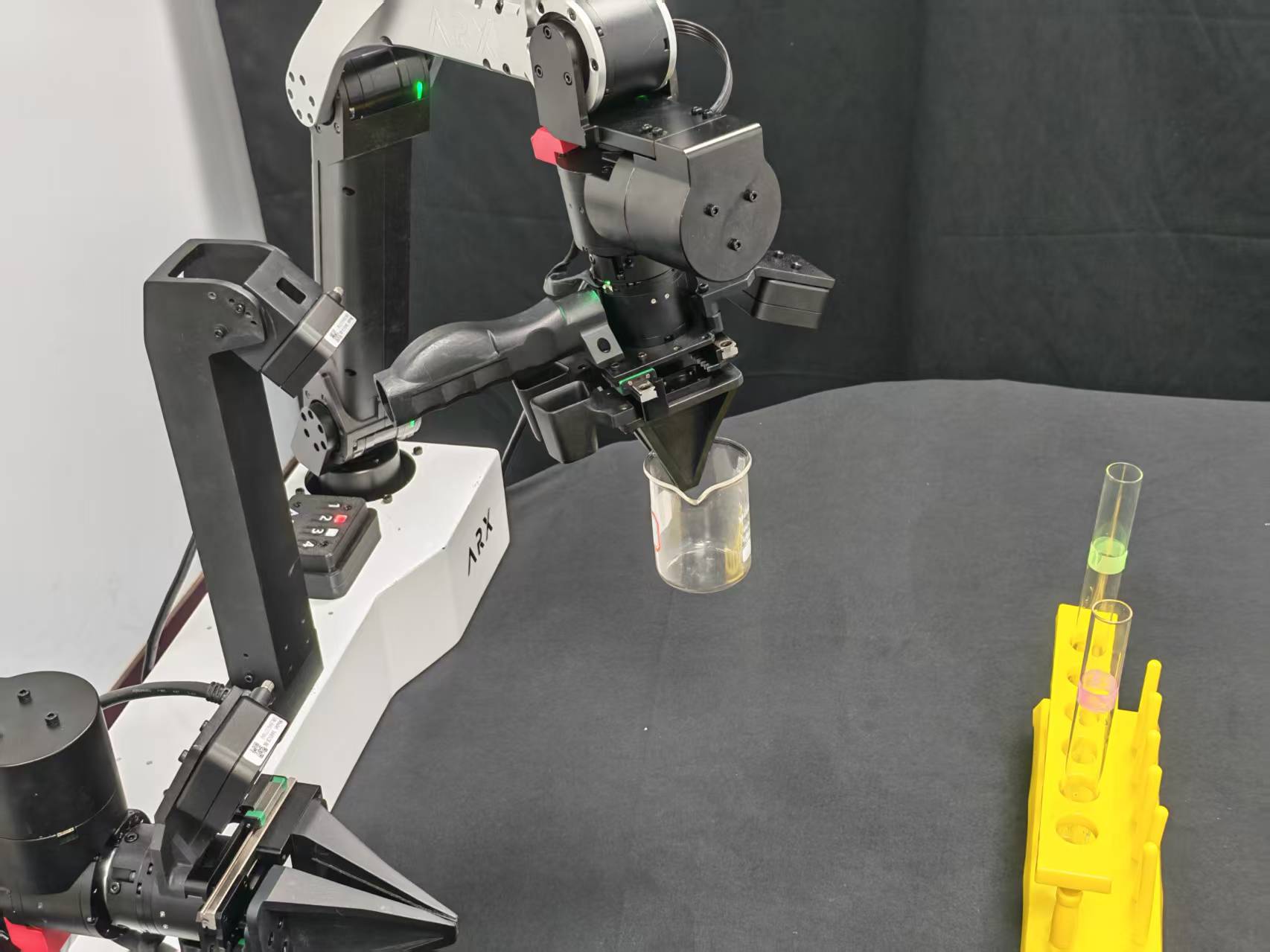} \\[0.8mm]
        {\scriptsize\textbf{(c)}} &
        \failedwideattempt{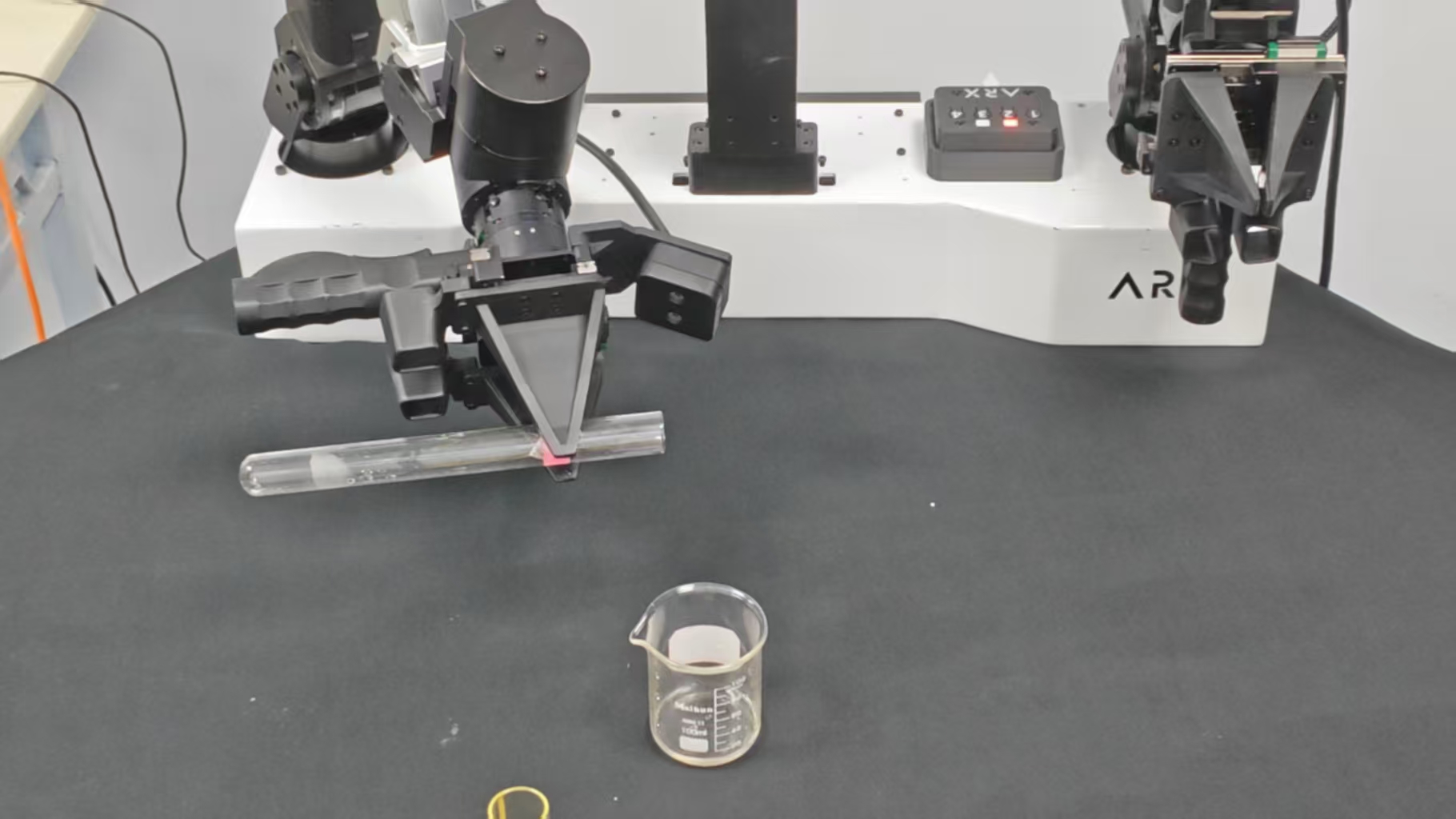} &
        \failedwideattempt{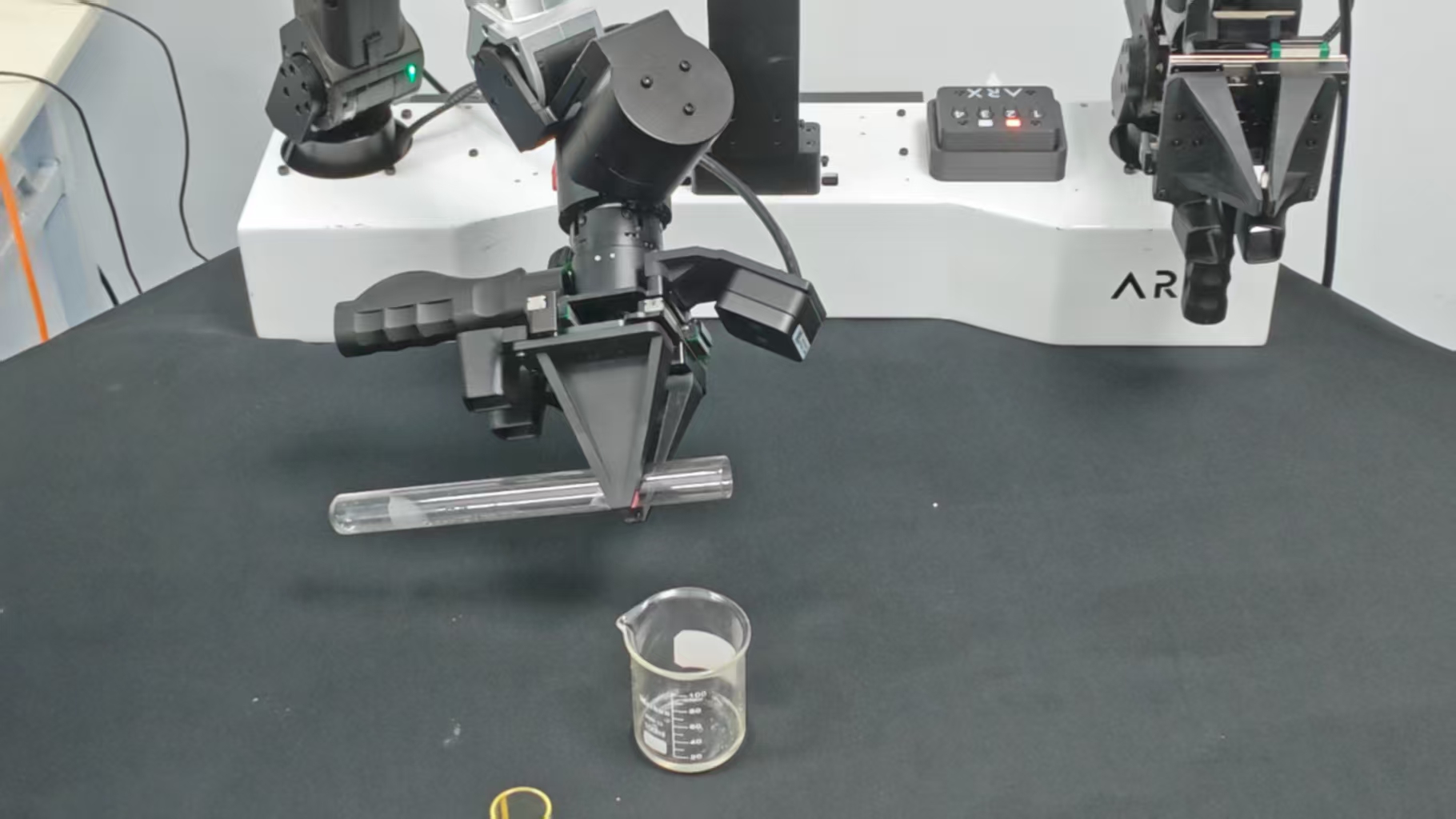} &
        \failedwideattempt{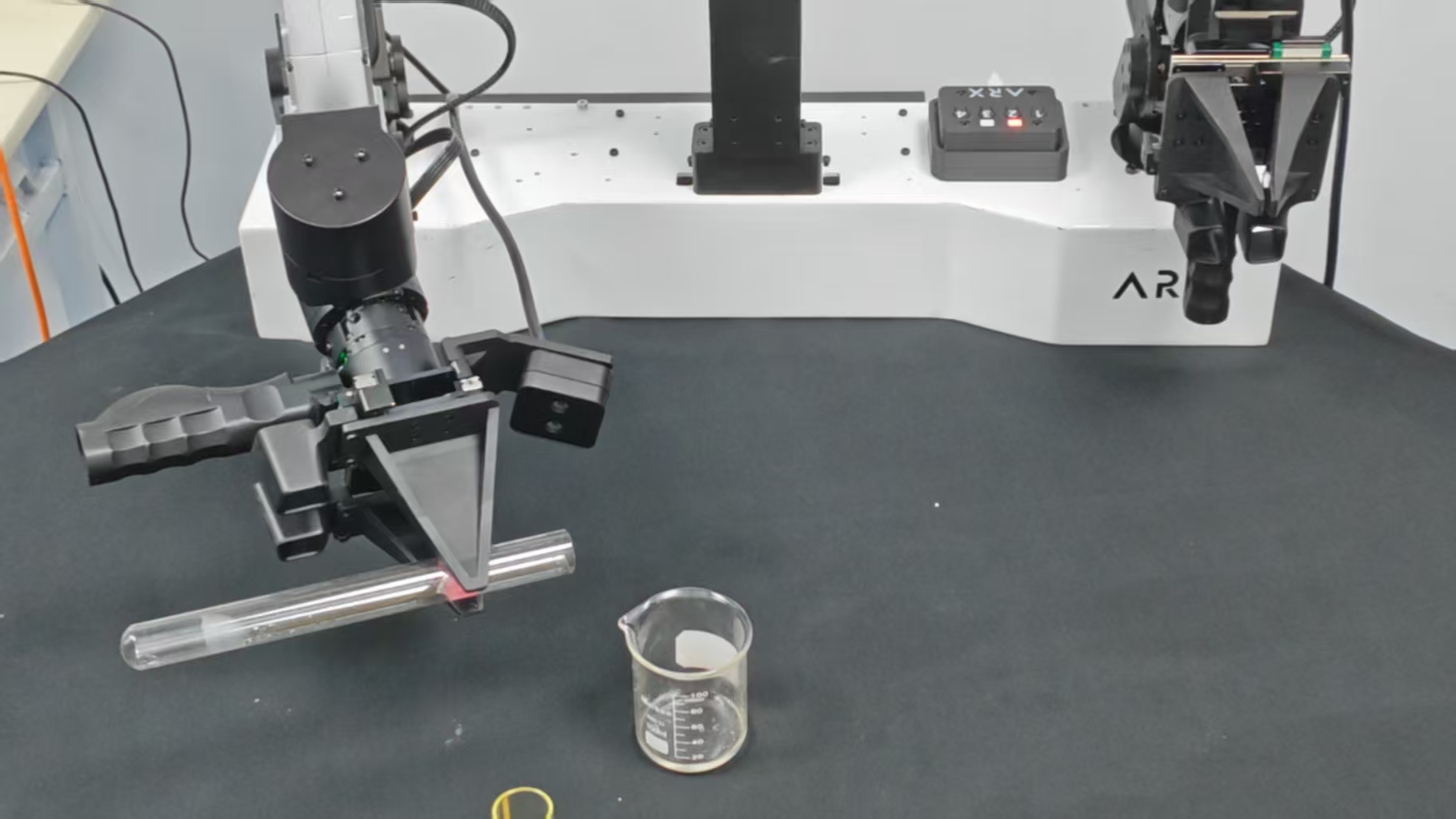} &
        \successfulwideattempt{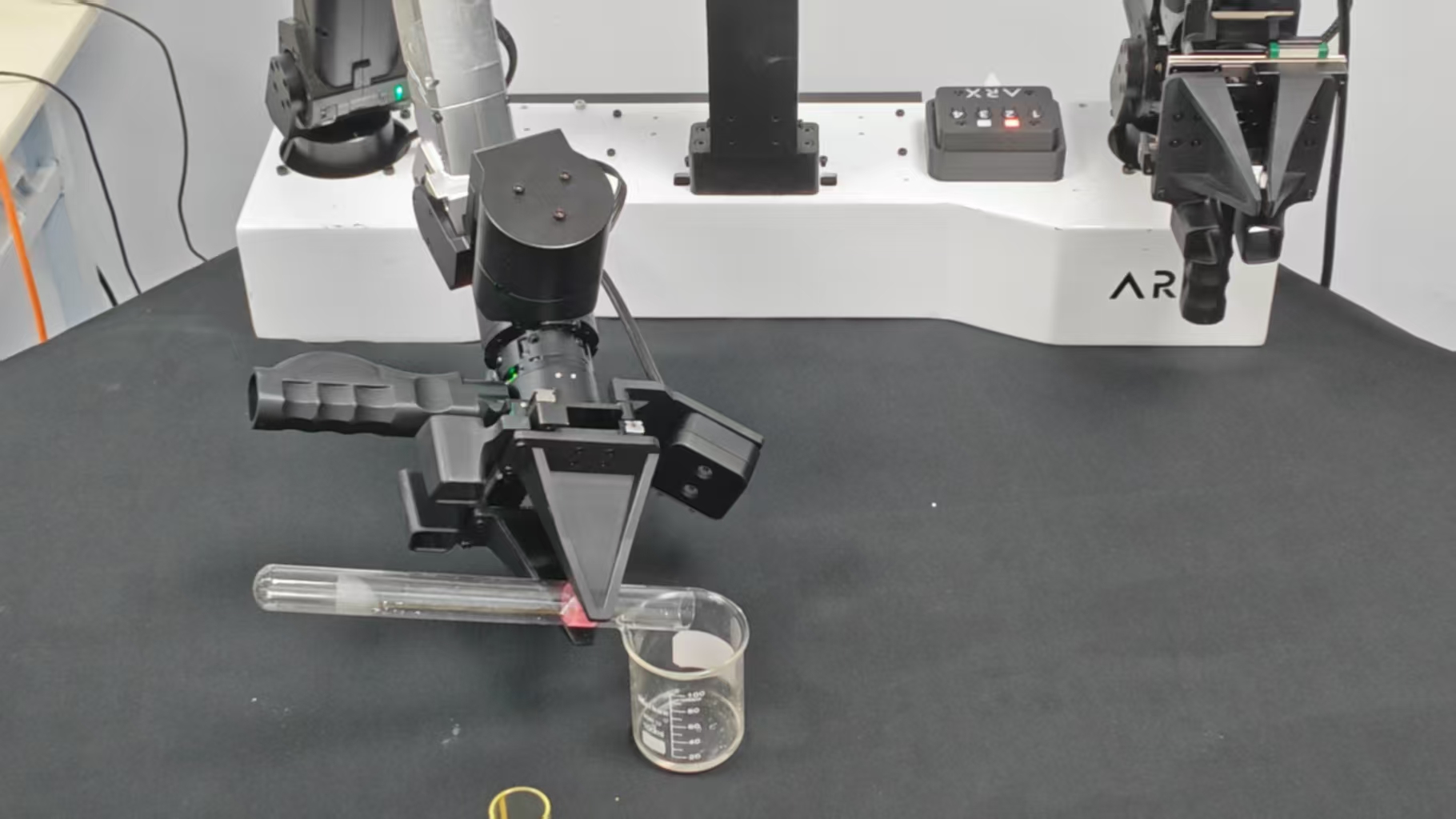}
    \end{tabular}
    \caption{Chronological post-deployment evolution on three ChemLab-Evo
    tasks. Red frames are terminal observations from distinct failed attempts.
    Green frames show the subsequent successful attempt; in (a) and (b), the
    two green frames are successive stages of the same continuous attempt.}
    \label{fig:qualitative-executions}
\end{figure}}

\newcommand{\HumanWarmupFigure}{%
\begin{figure*}[!t]
    \centering
    \begin{minipage}[c]{0.66\textwidth}
        \centering
        \setlength{\tabcolsep}{0.6pt}
        \resizebox{\linewidth}{!}{%
        \begin{tabular}{@{}rcccc@{}}
            {\scriptsize\textbf{(a)}} &
            \includegraphics[width=0.18\linewidth]{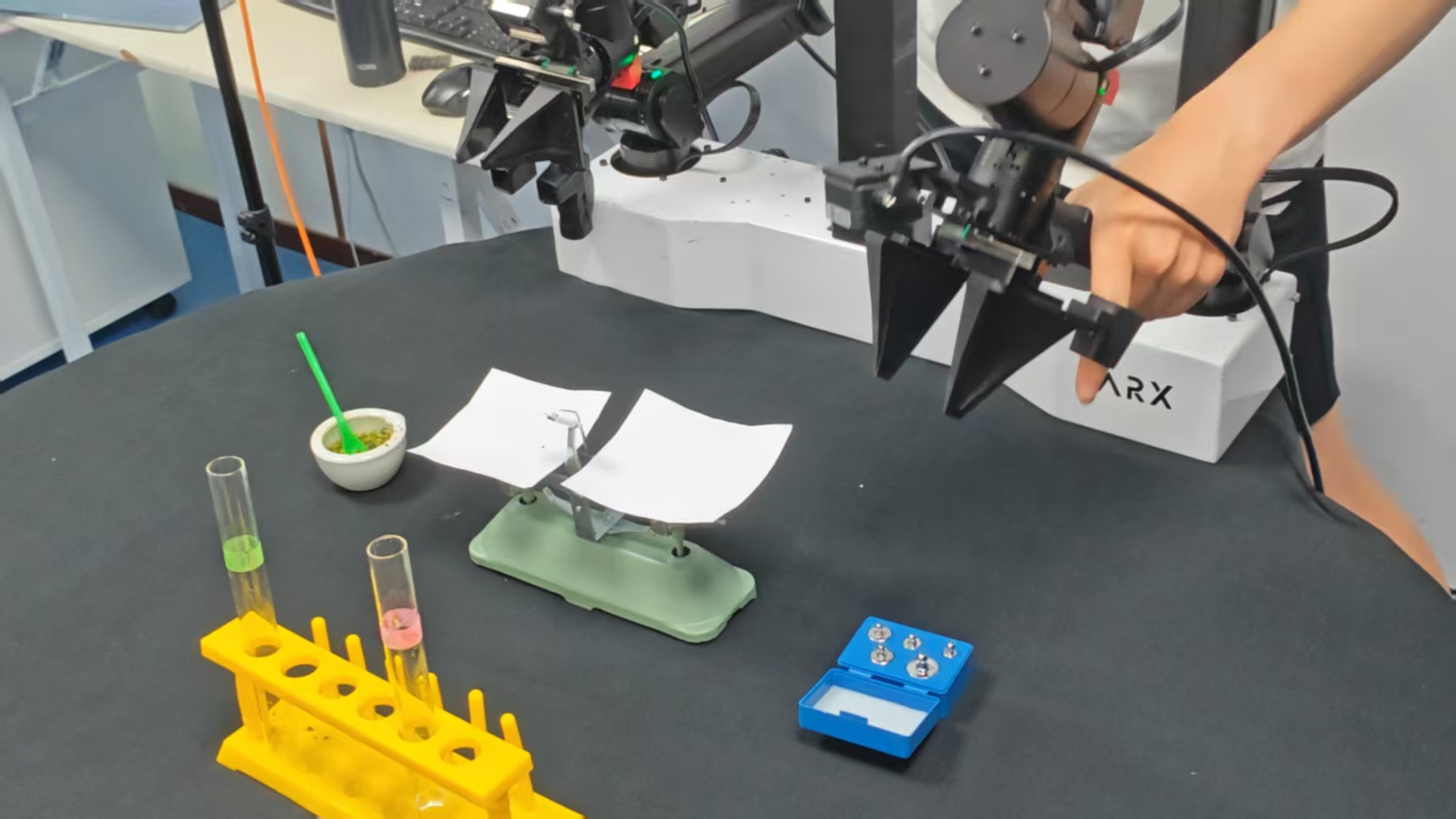} &
            \includegraphics[width=0.18\linewidth]{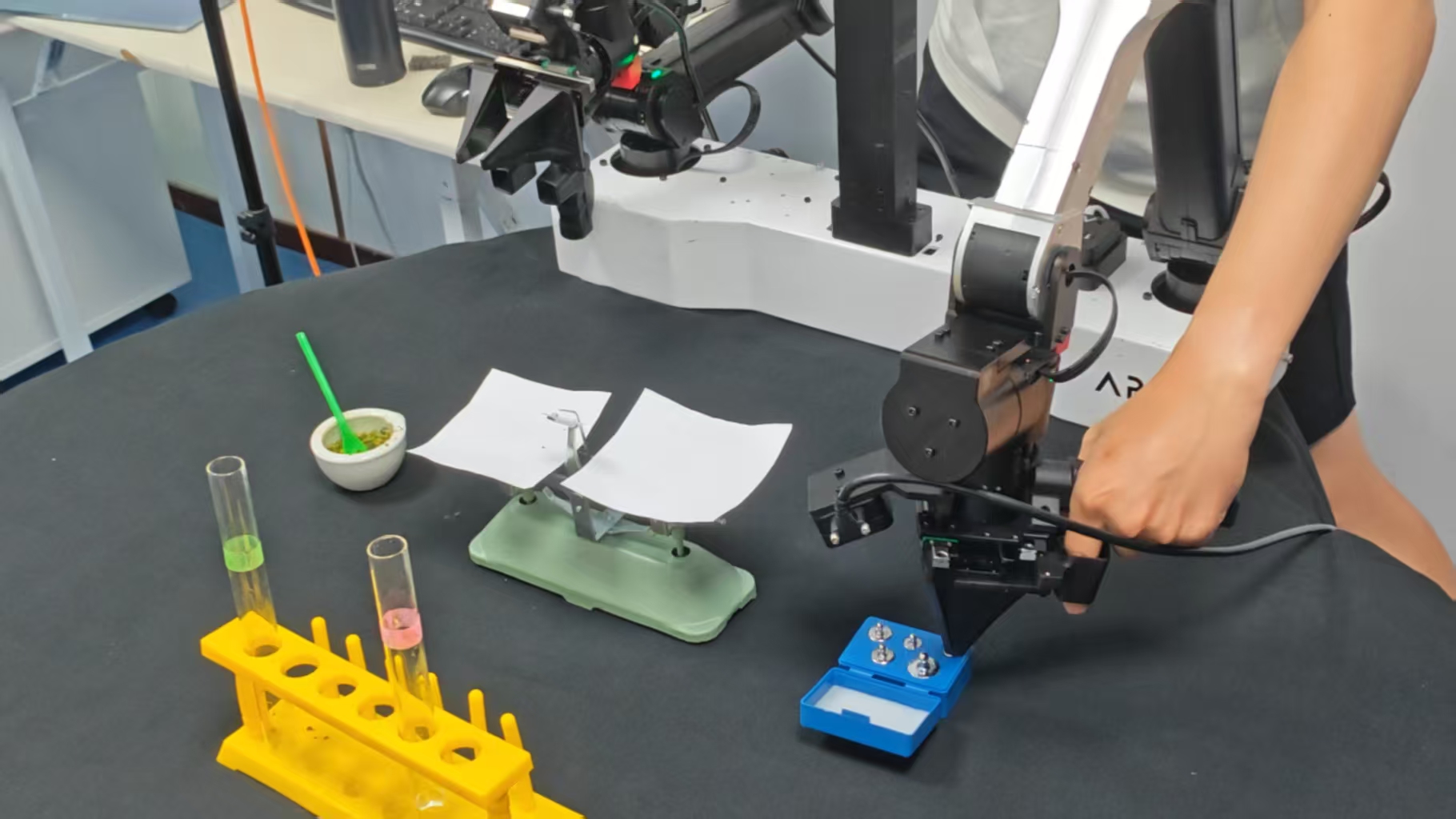} &
            \includegraphics[width=0.18\linewidth]{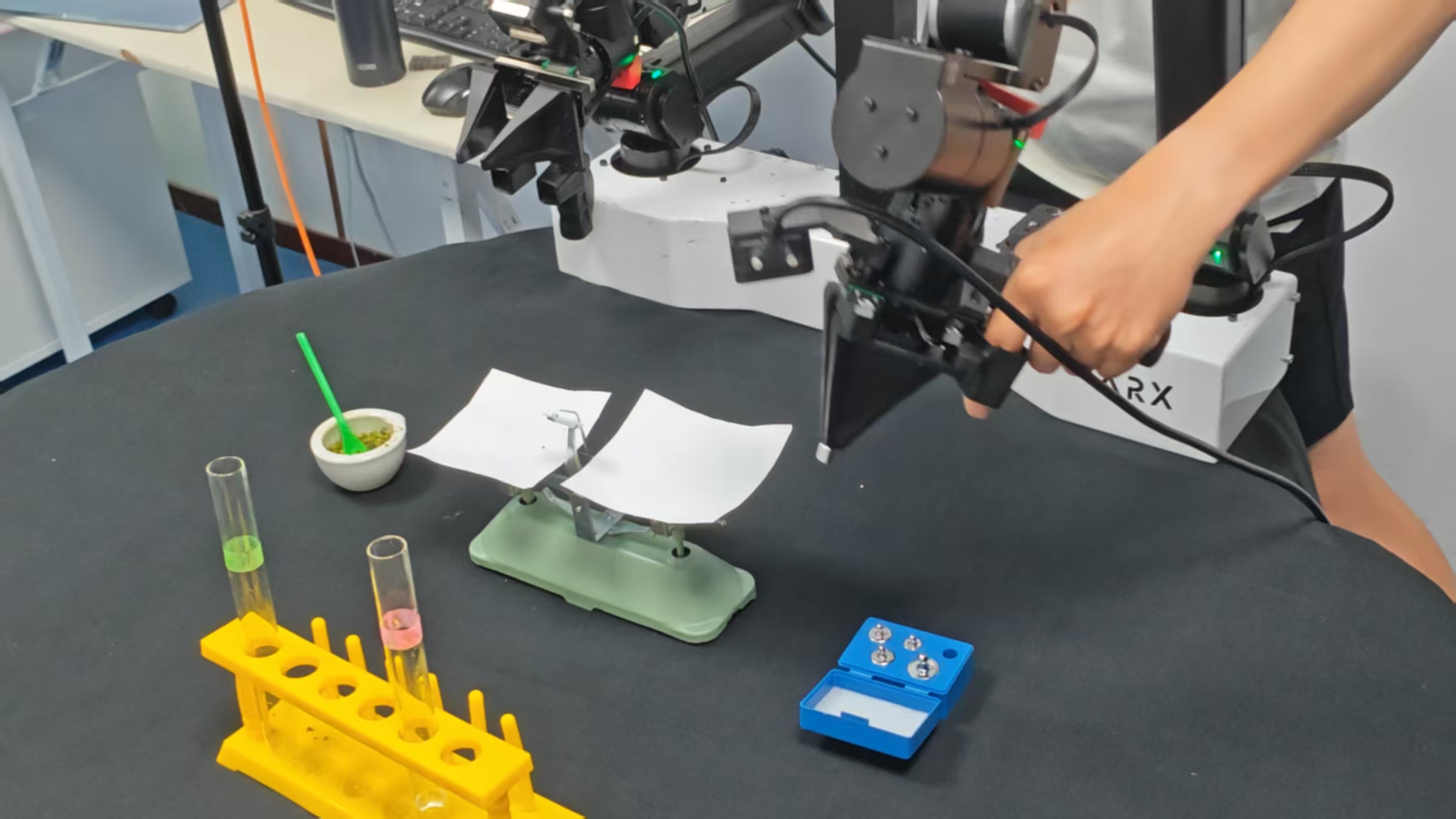} &
            \includegraphics[width=0.18\linewidth]{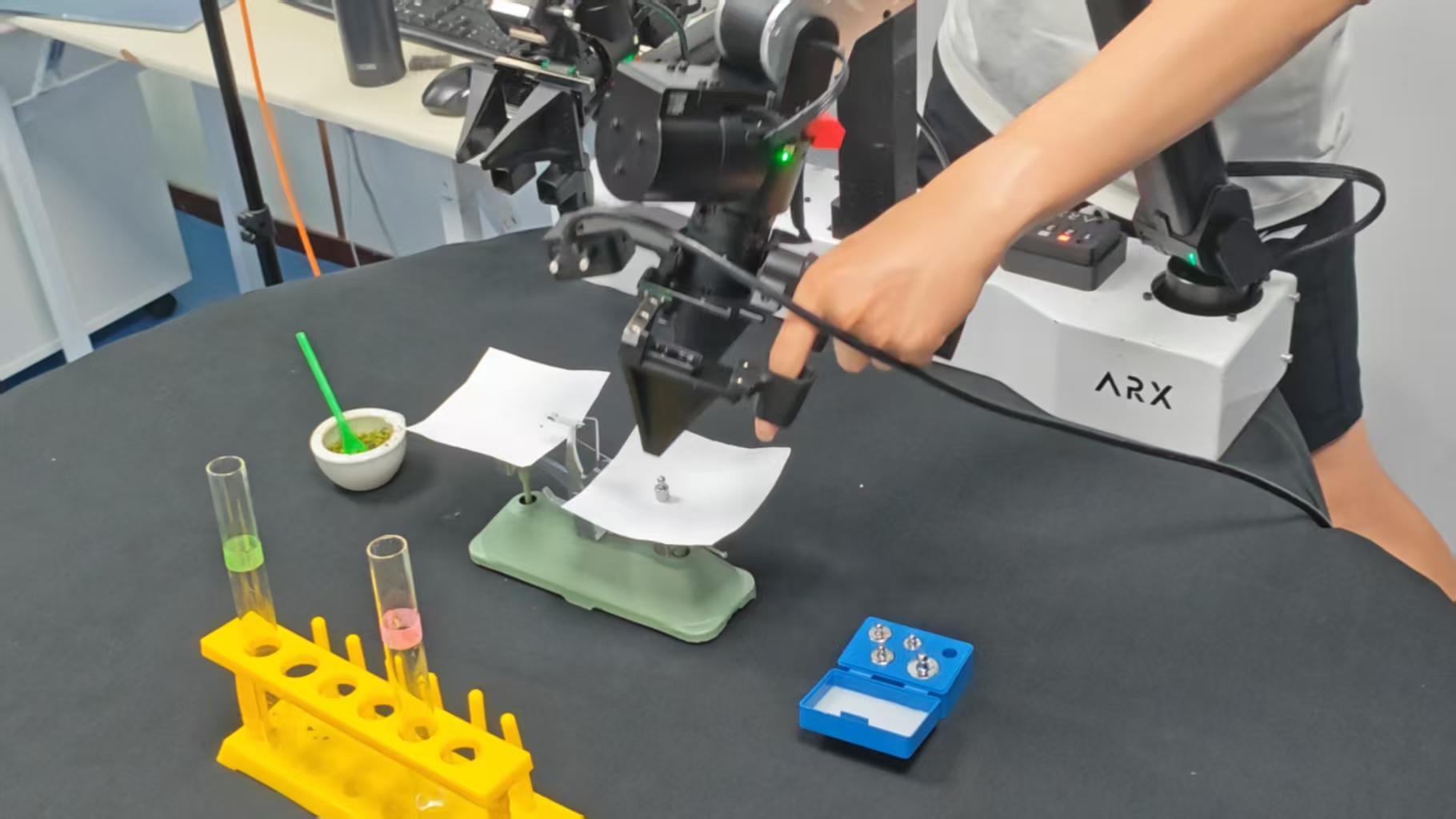} \\[0.8mm]
            {\scriptsize\textbf{(b)}} &
            \includegraphics[width=0.18\linewidth]{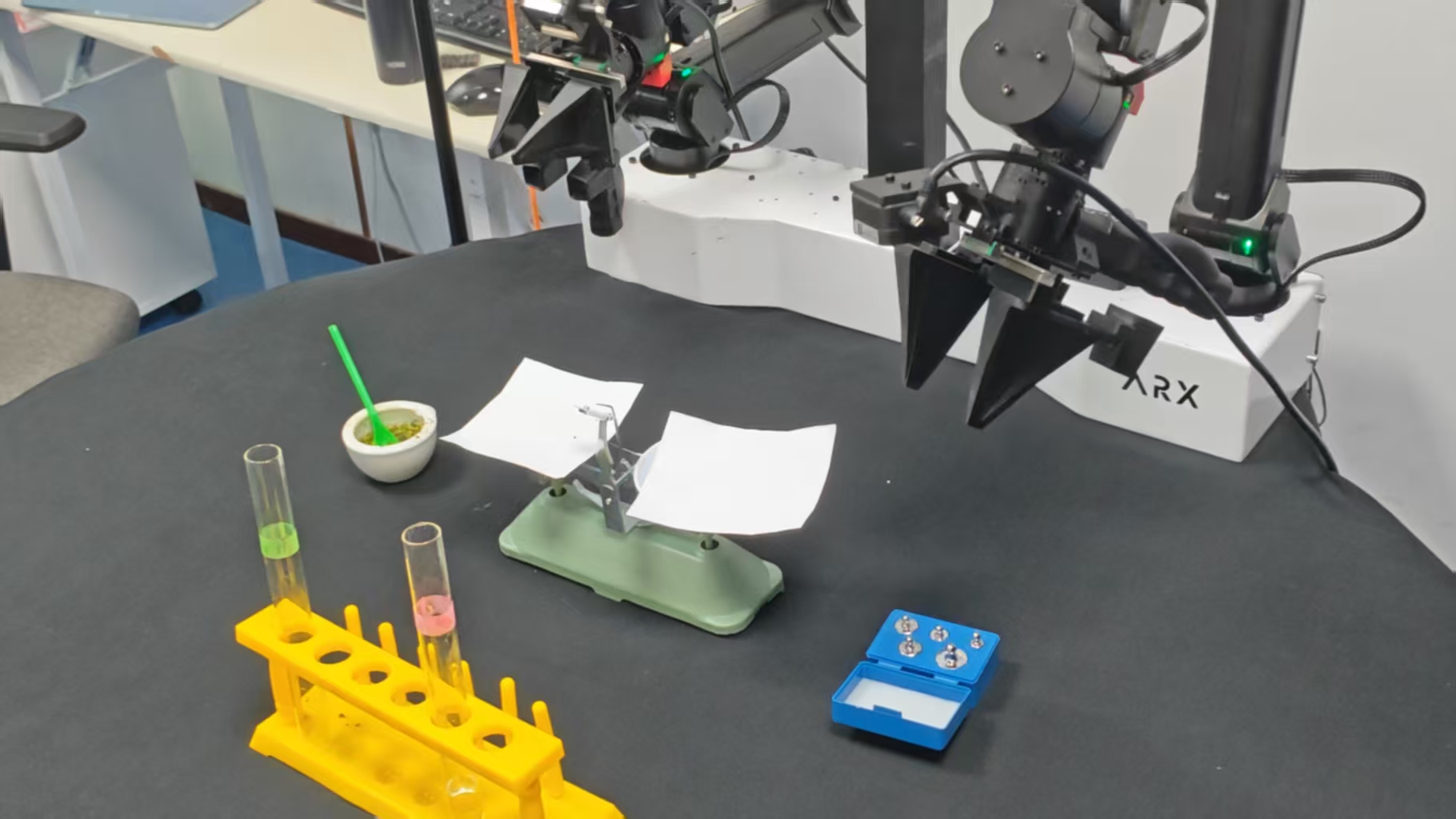} &
            \includegraphics[width=0.18\linewidth]{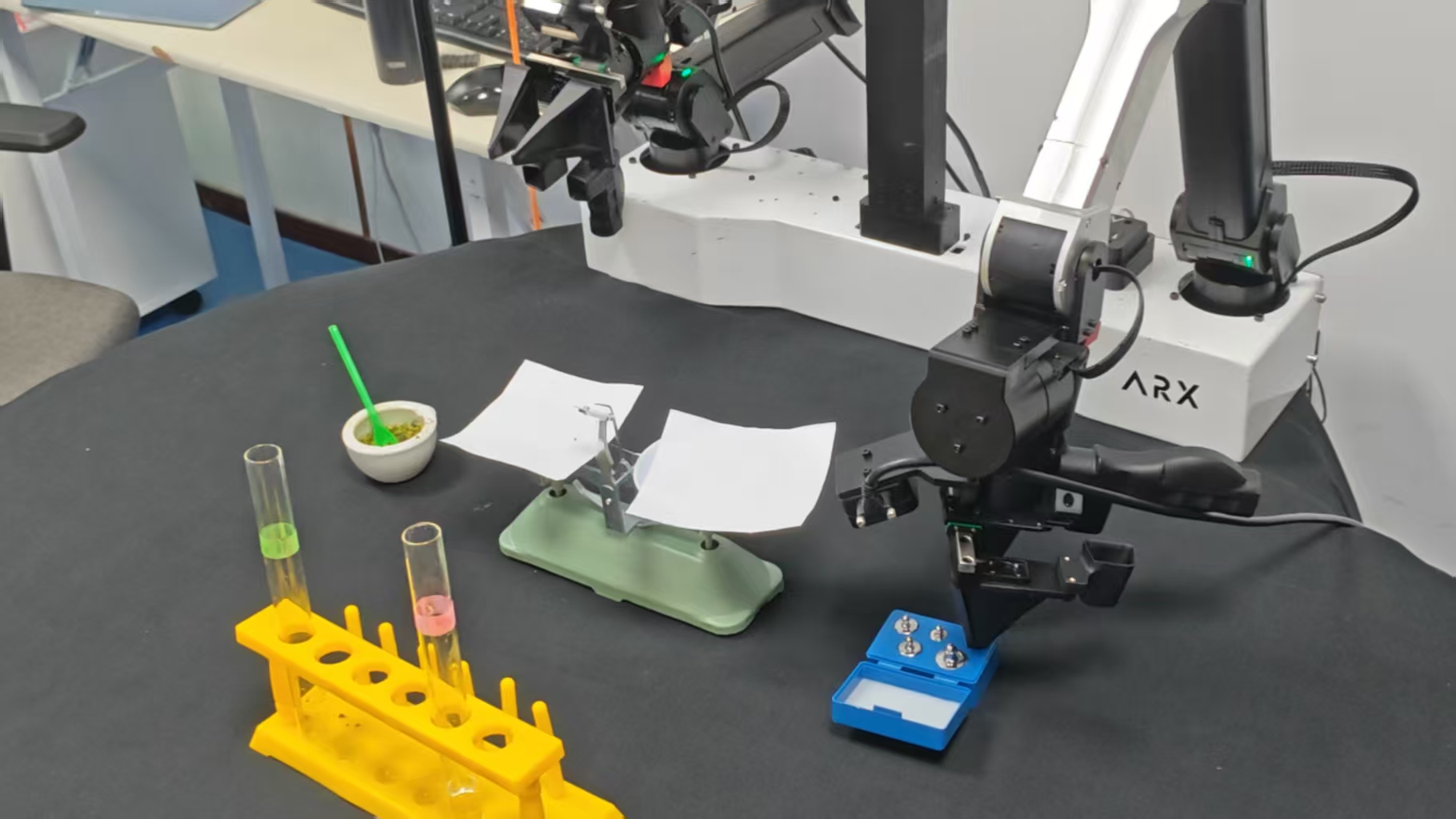} &
            \includegraphics[width=0.18\linewidth]{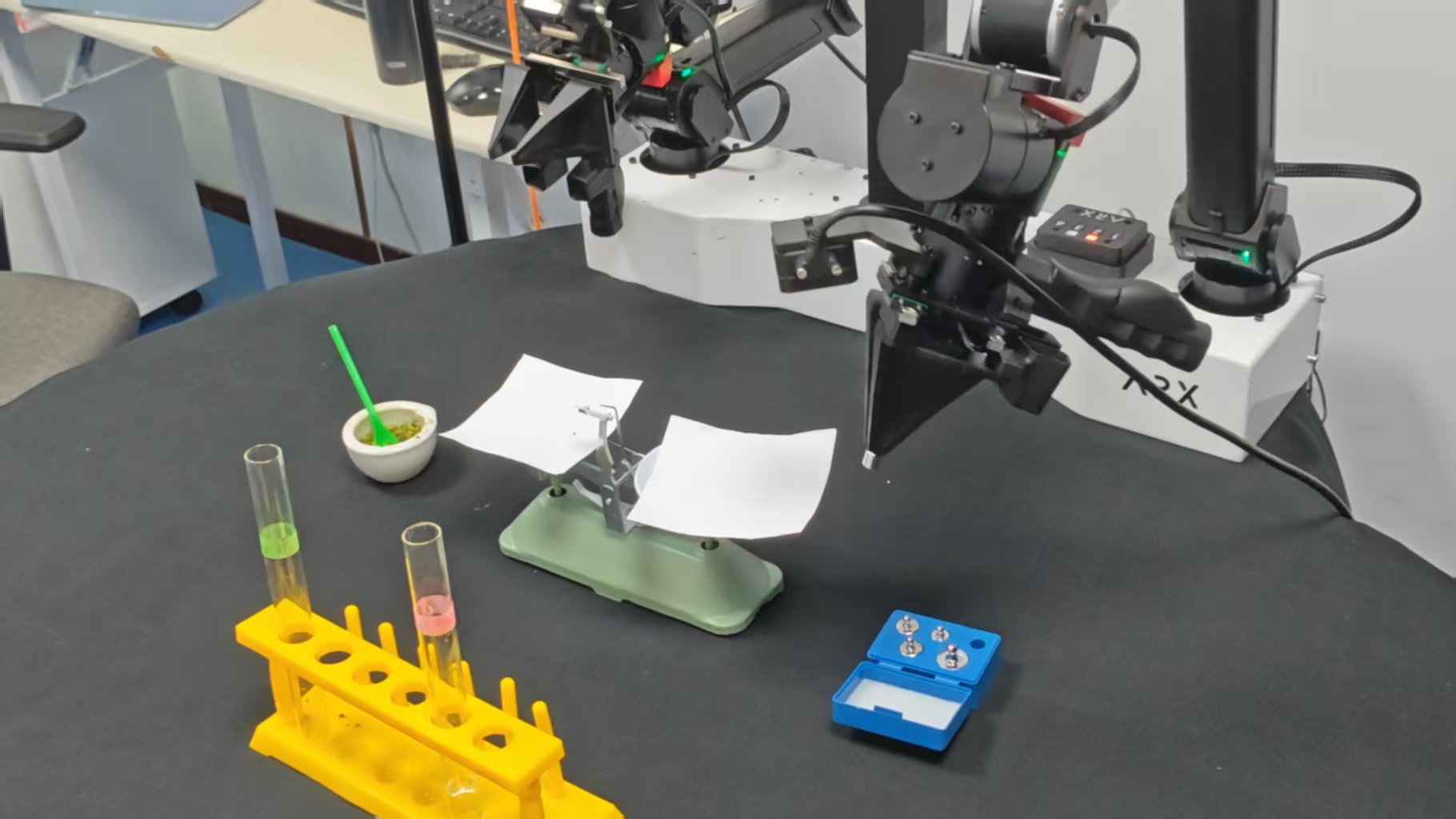} &
            \includegraphics[width=0.18\linewidth]{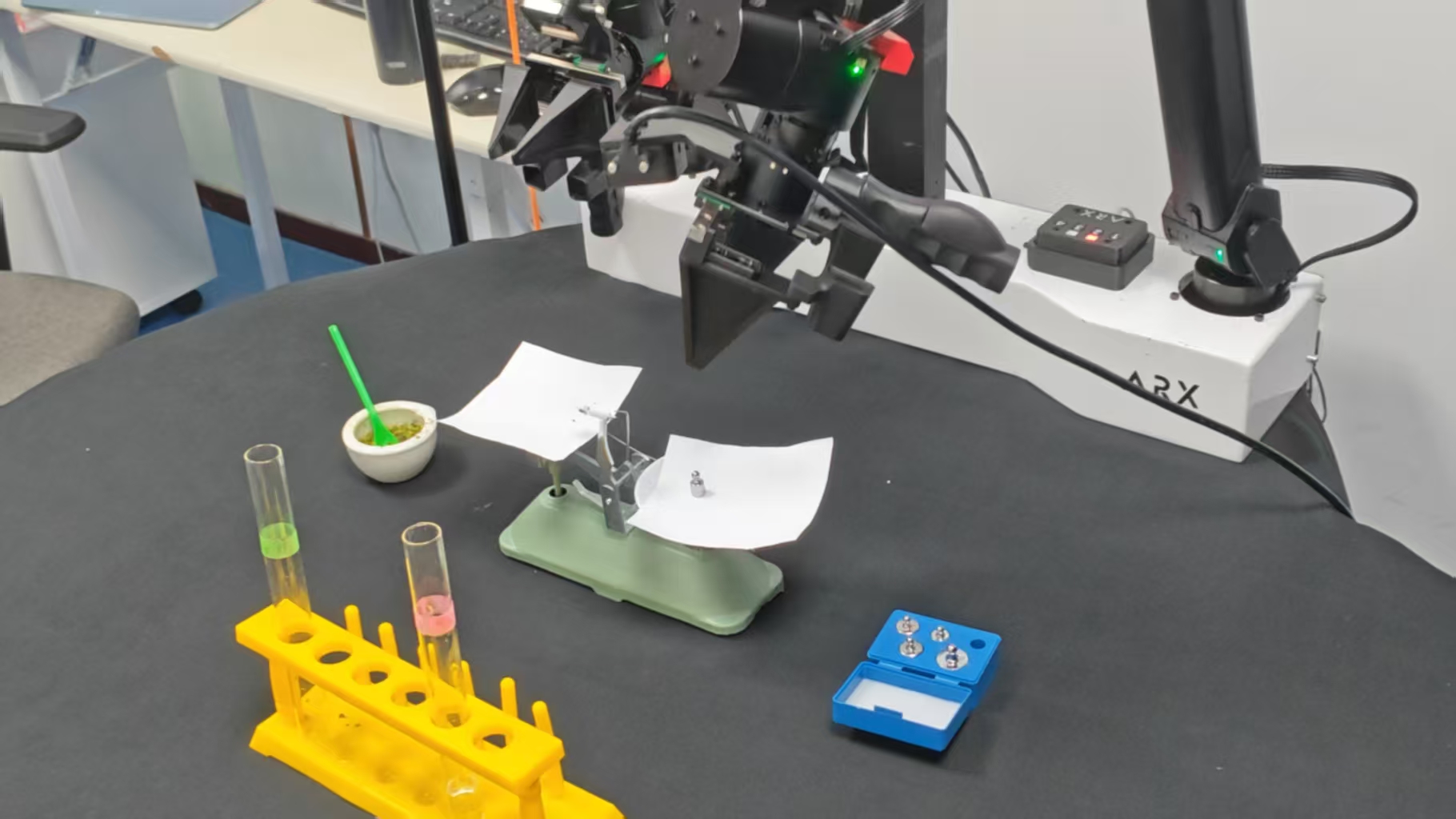} \\[0.8mm]
            {\scriptsize\textbf{(c)}} &
            \includegraphics[width=0.18\linewidth]{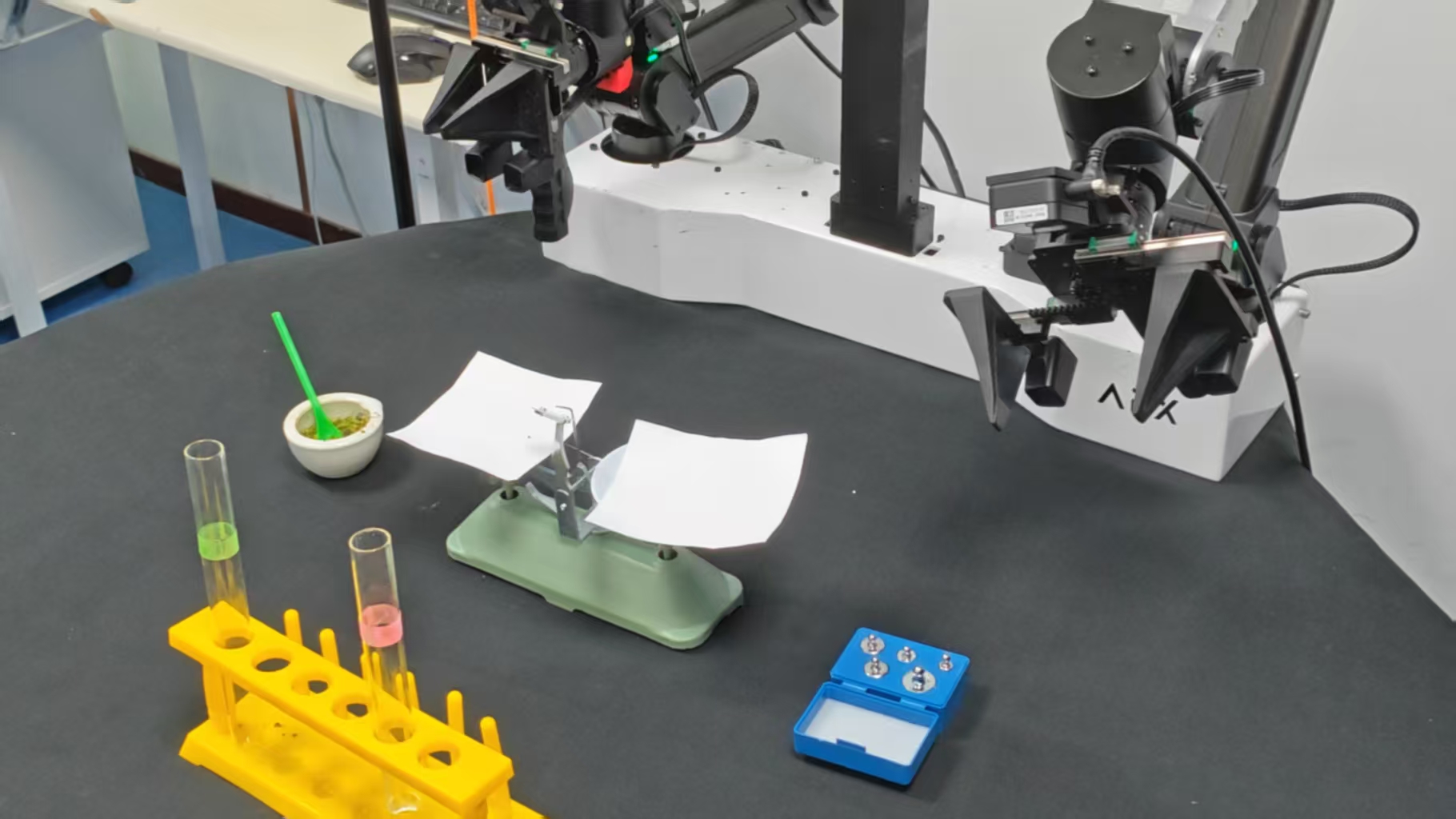} &
            \includegraphics[width=0.18\linewidth]{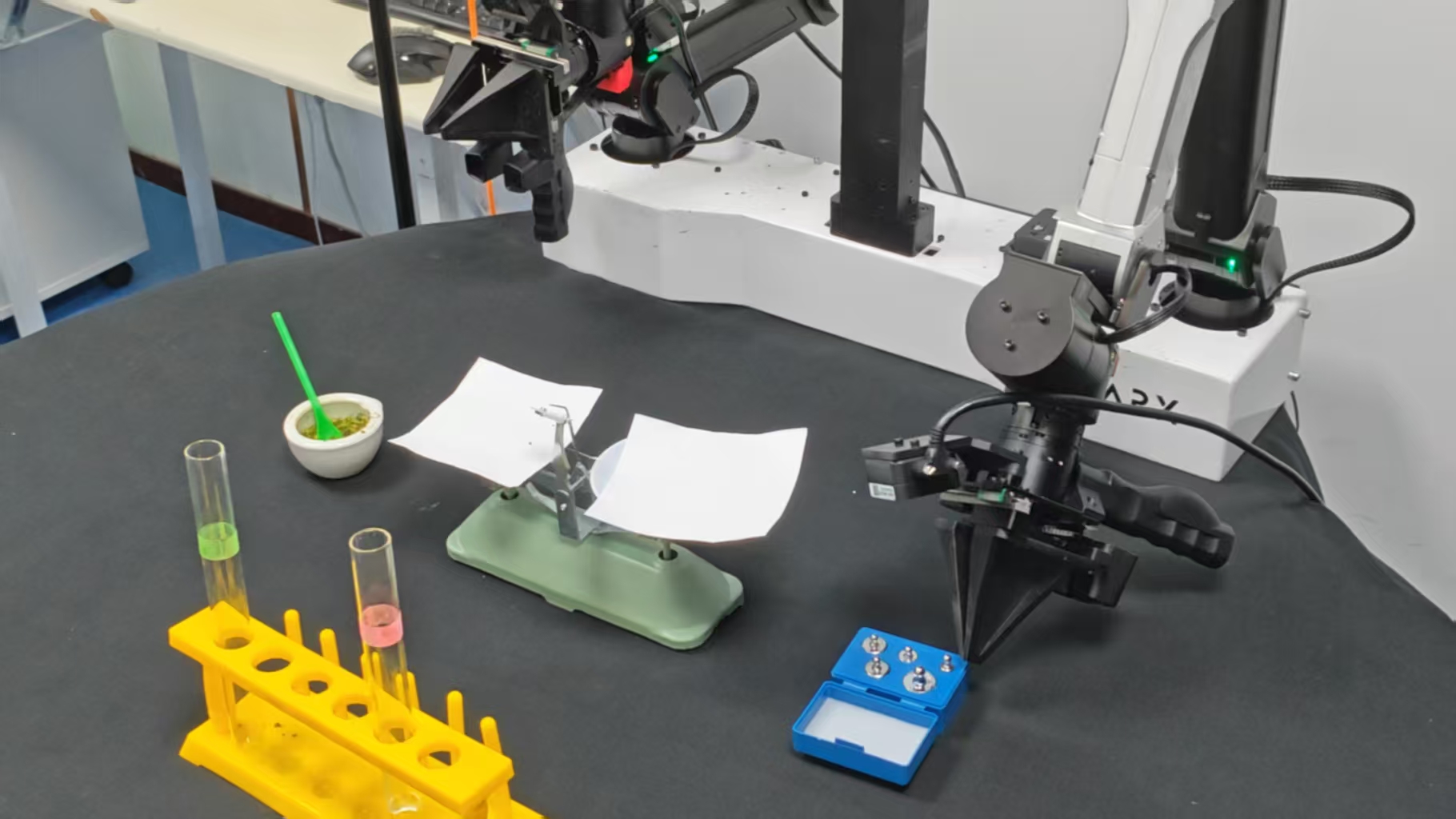} &
            \includegraphics[width=0.18\linewidth]{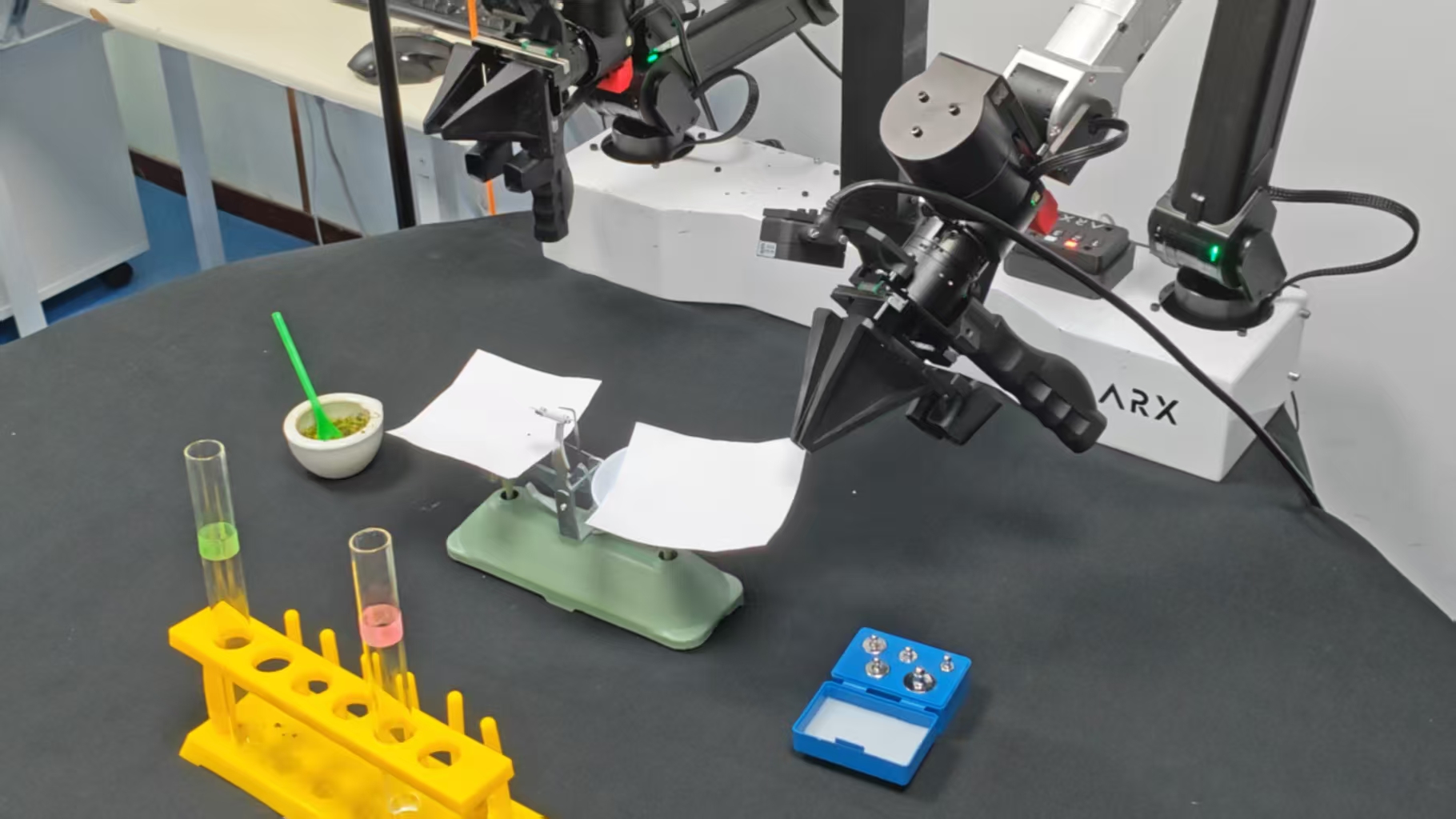} &
            \includegraphics[width=0.18\linewidth]{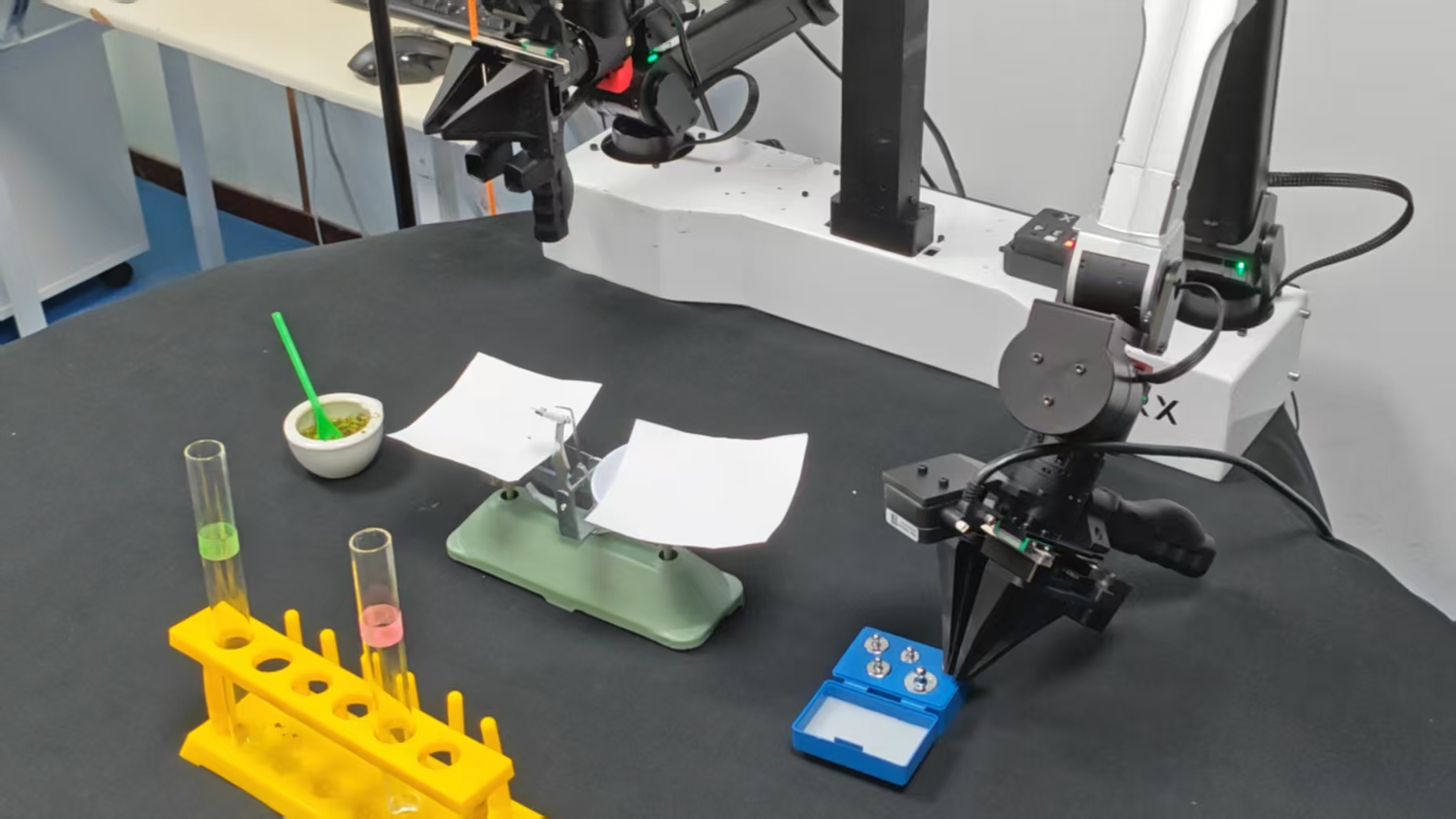}
        \end{tabular}}
    \end{minipage}\hfill
    \begin{minipage}[c]{0.31\textwidth}
        \centering
        \includegraphics[width=0.94\linewidth,trim=145 90 100 120,clip]
            {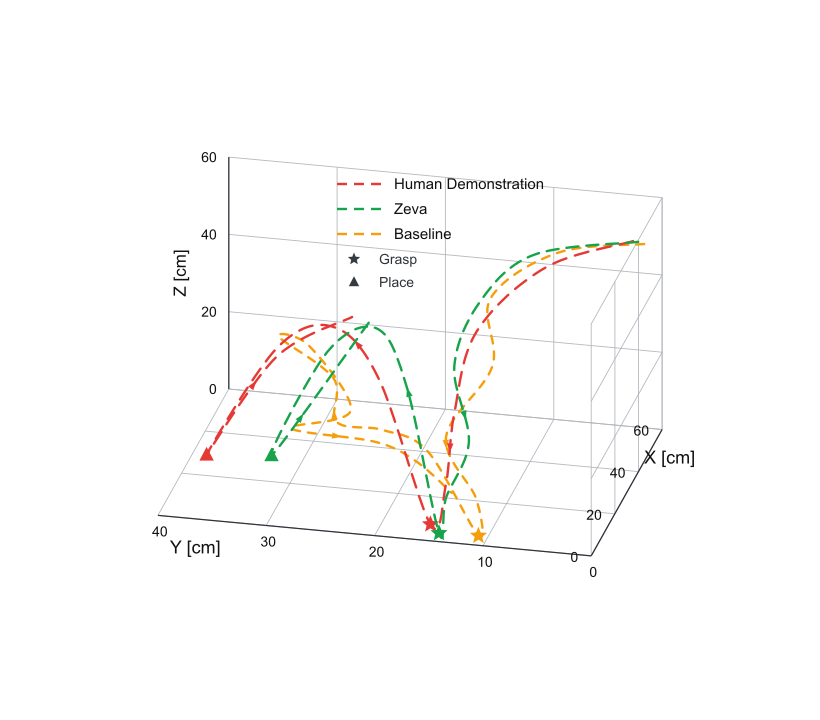}
    \end{minipage}
    \caption{One-shot human warm-up for Balance Weighing. Rows show
    (a) the human-guided demonstration used to initialize PIM,
    (b) Zeva after one-shot memory warm-up, and
    (c) the baseline without memory warm-up. Frames in each row form
    a timeline ordered from left to right. The 3D plot compares the corresponding
    end-effector trajectories; stars and triangles denote grasp and placement
    events, respectively.}
    \label{fig:human-warmup}
\end{figure*}}

\newcommand{\HumanWarmupScalingFigure}{%
\begin{figure*}[!t]
    \centering
    \includegraphics[width=0.98\textwidth]{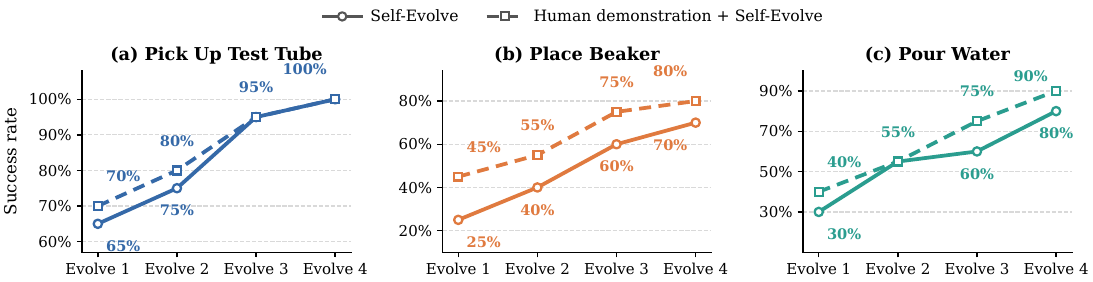}
    \caption{One-shot human-demonstration warm-up on three atomic ChemLab-Evo
    tasks. Each evaluated fixed episode is paired with one task-matched human
    demonstration before autonomous self-evolution. Solid lines show
    self-evolution without warm-up, while dashed lines show human warm-up
    followed by self-evolution. Each
    panel uses its own y-axis range to keep within-task changes legible.}
    \label{fig:human-warmup-scaling}
\end{figure*}}

\newcommand{\effectpairwide}[3]{%
    \begin{minipage}[c]{0.285\textwidth}
        \centering
        \begingroup
        \setlength{\fboxsep}{1.0pt}
        \setlength{\fboxrule}{0.8pt}
        \fcolorbox{#3}{white}{%
            \begin{tabular}{@{}c@{\hspace{0.8mm}}c@{\hspace{0.8mm}}c@{}}
                \includegraphics[width=0.41\linewidth]{#1} &
                {\color{gray!70}\large$\rightarrow$} &
                \includegraphics[width=0.41\linewidth]{#2}
            \end{tabular}}
        \endgroup
    \end{minipage}}

\newcommand{\CrossTaskEffectSection}{%
\begin{figure*}[!t]
    \centering
    \setlength{\tabcolsep}{1.0pt}
    \begin{tabular}{@{}>{\raggedright\arraybackslash}m{0.105\textwidth}ccc@{}}
        \small\textbf{Pouring} &
        \effectpairwide{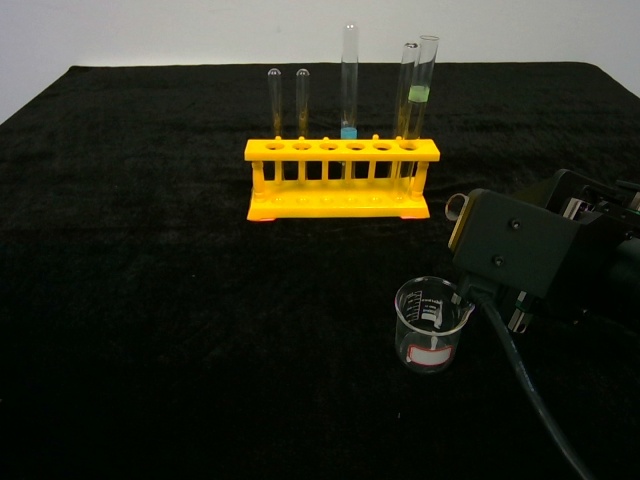}{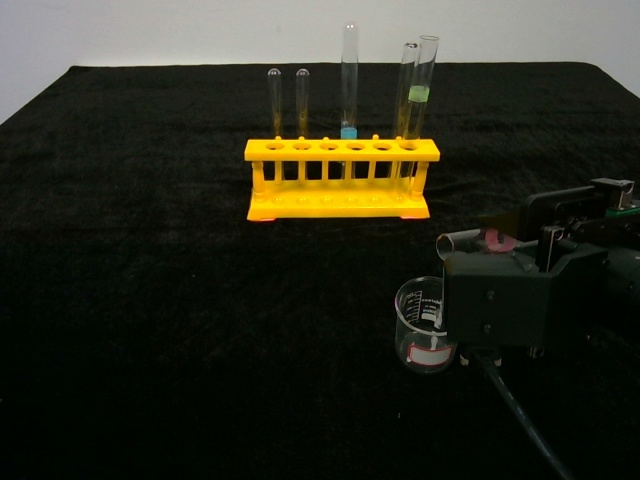}{blue!70!black} &
        \effectpairwide{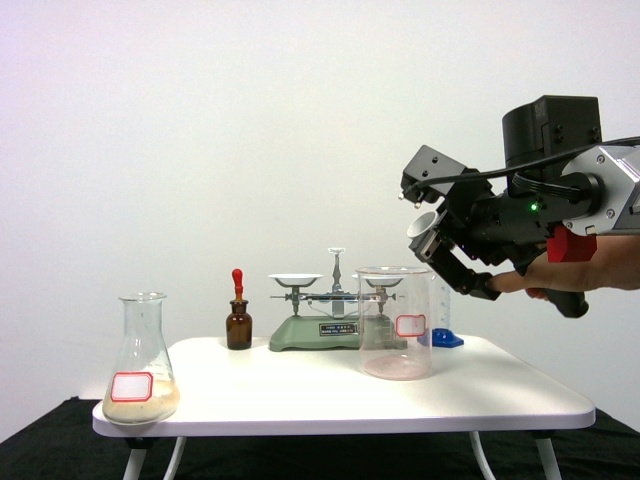}{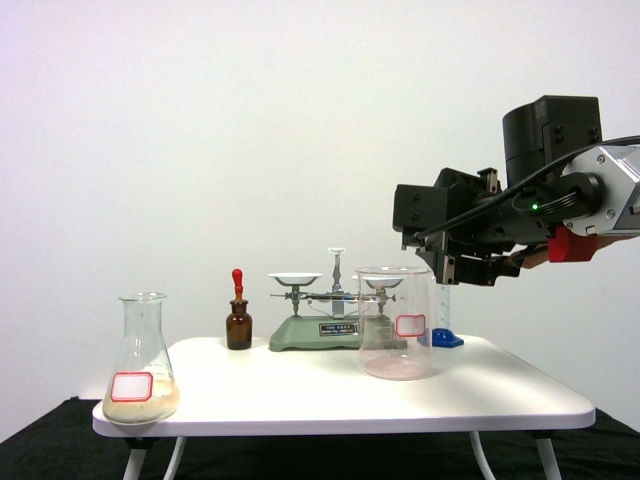}{teal!70!black} &
        \effectpairwide{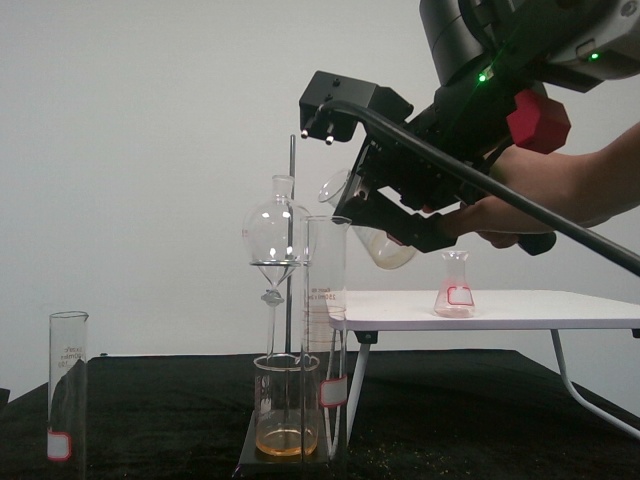}{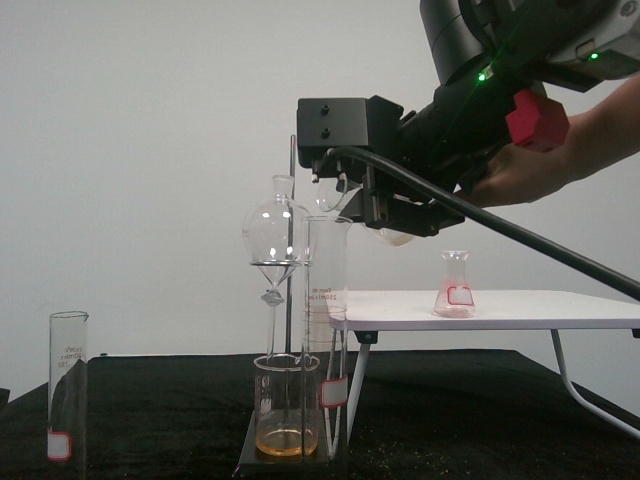}{teal!70!black} \\[2.0mm]
        \small\textbf{Closing} &
        \effectpairwide{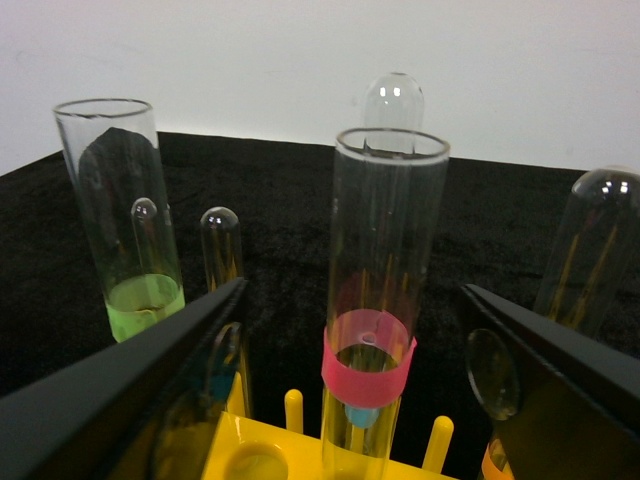}{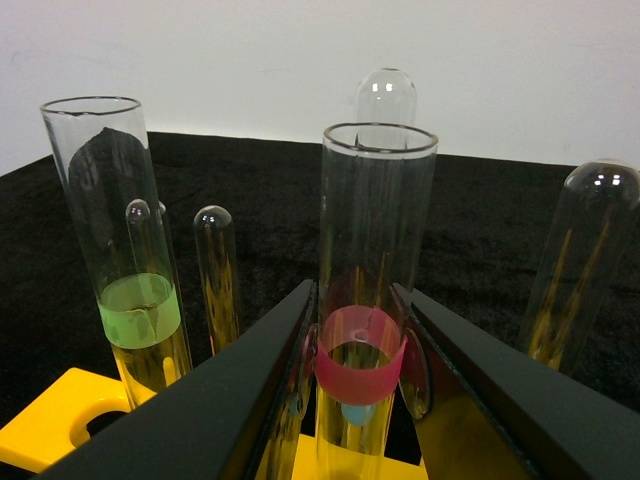}{blue!70!black} &
        \effectpairwide{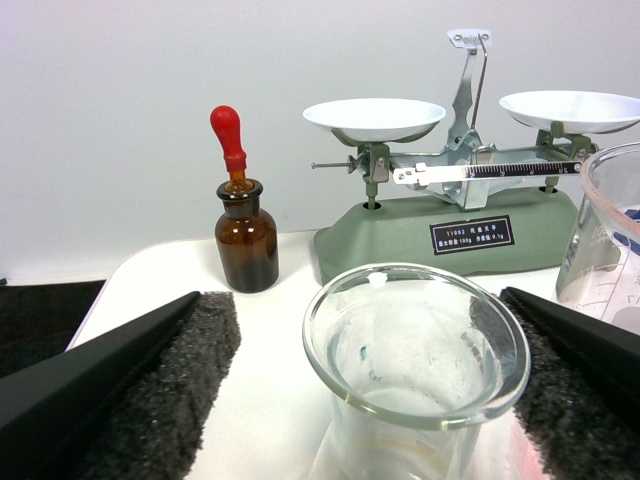}{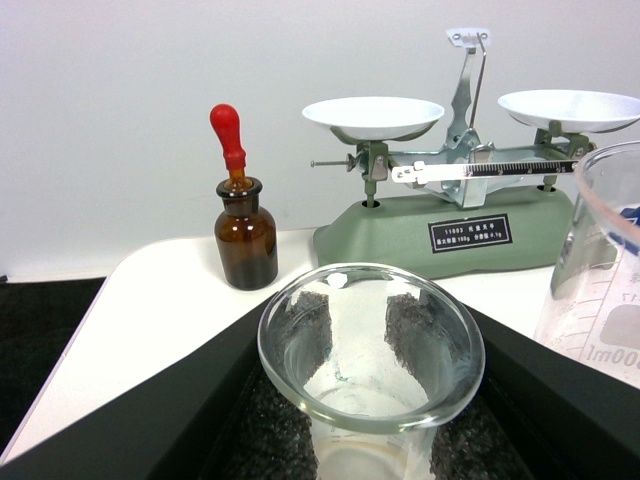}{teal!70!black} &
        \effectpairwide{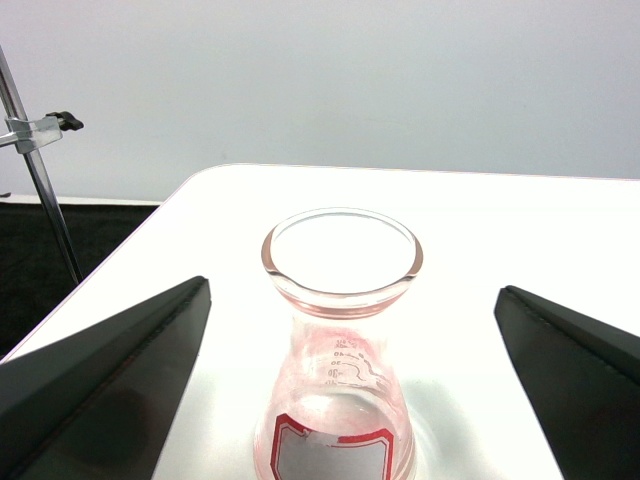}{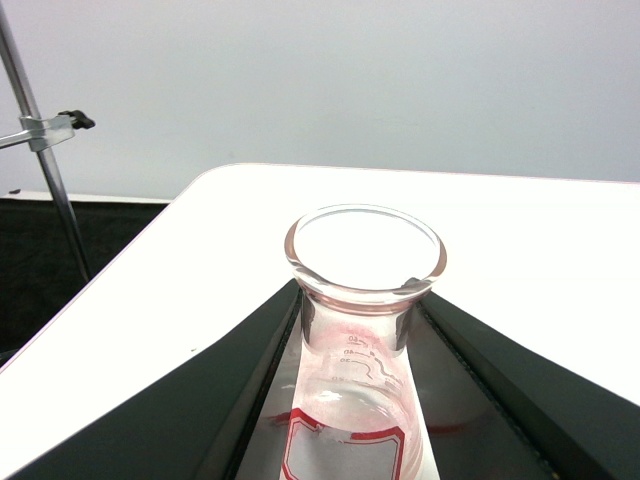}{teal!70!black} \\[2.0mm]
        \small\textbf{Lifting} &
        \effectpairwide{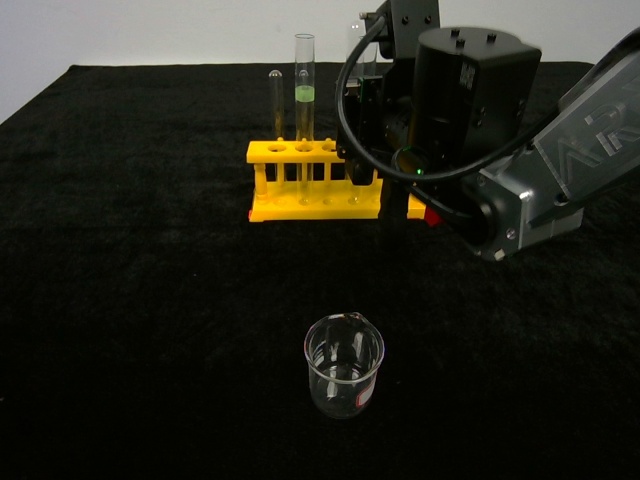}{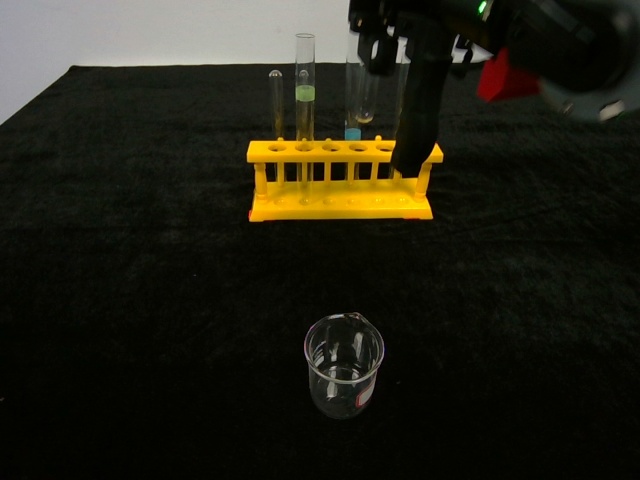}{blue!70!black} &
        \effectpairwide{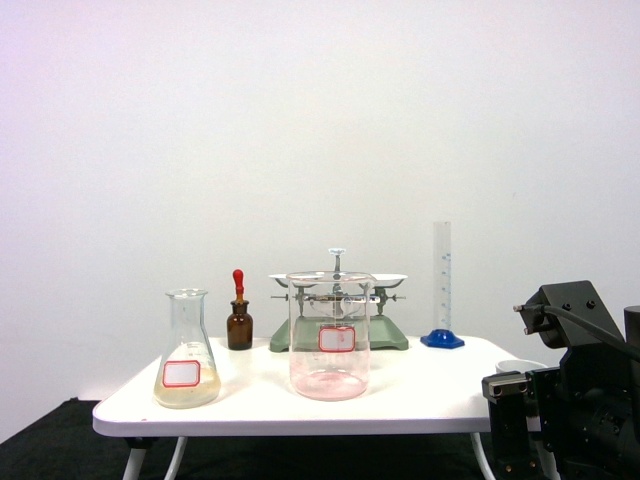}{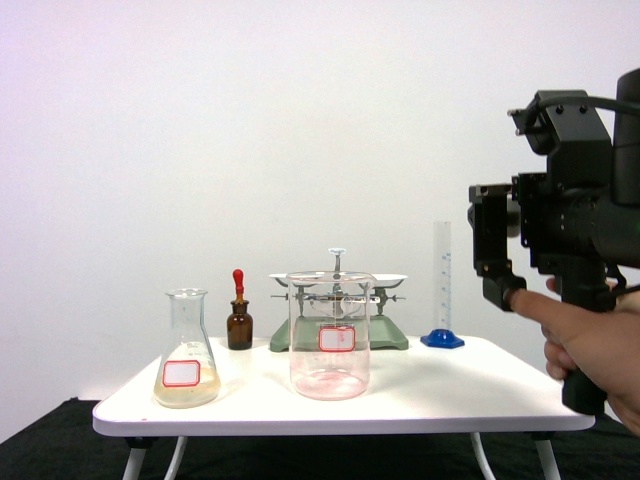}{teal!70!black} &
        \effectpairwide{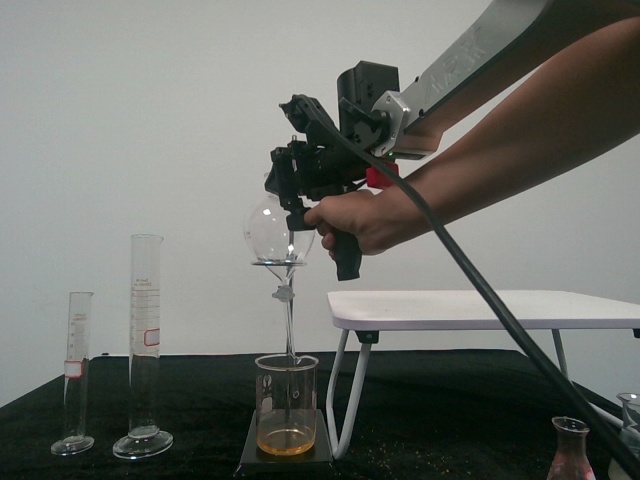}{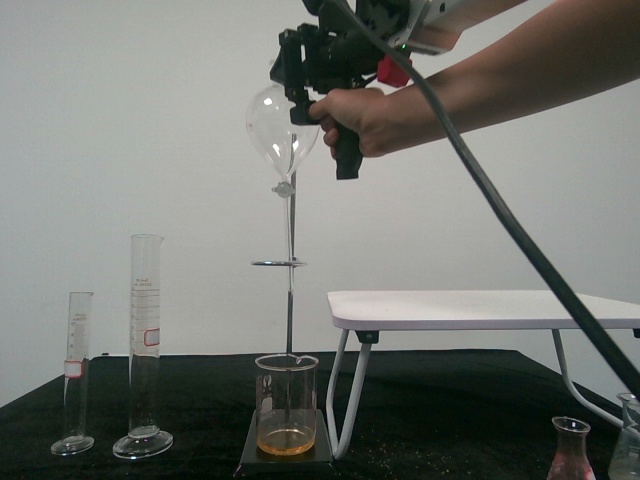}{teal!70!black}
    \end{tabular}
    \caption{Cross-task retrieval in the latent space of CTE causal interaction
    signals.
    Each block shows a left-to-right state transition. Blue outlines denote
    queries, and teal outlines denote their two nearest cross-task matches.
    Rows group transitions by physical effect despite changes in objects,
    viewpoints, and instructions.}
    \label{fig:cross-task-effect-retrieval}
\end{figure*}}
\newcommand{\CrossTaskEffectTransferFigure}{%
\begin{figure}[H]
    \centering
    \includegraphics[width=0.92\linewidth]{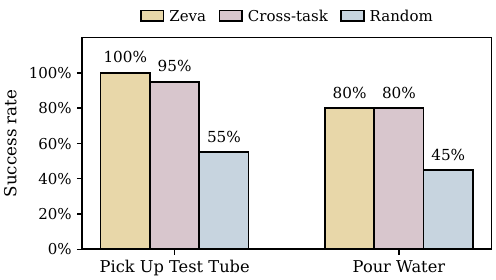}
    \caption{Quantitative replacement with cross-task causal interaction
    signals on ChemLab-Evo. Zeva uses the retrieved task-local interaction signal,
    Cross-task substitutes its nearest counterpart from another task, and
    Random uses a randomly selected interaction signal from the same cross-task
    pool. The frozen policy and all other inputs remain unchanged.}
    \label{fig:cross-task-effect-transfer}
\end{figure}}

\def\BibTeX{{\rm B\kern-.05em{\sc i\kern-.025em b}\kern-.08em
    T\kern-.1667em\lower.7ex\hbox{E}\kern-.125emX}}

\begin{document}

\title{\name: In-Context Causal Learning for Generalizable Embodied Manipulation}

\author{
\IEEEauthorblockN{
Fu Chen\textsuperscript{*},
Xin Ding\textsuperscript{*,\dag},
Bingjia Huang,
Xiangyu Li,
Mingju Wang,
Jiawei He\\
Kun Li,
Wei Sun,
Yunxin Liu,
Hao Wu\textsuperscript{\dag,\ddag},
Ting Cao\textsuperscript{\dag,\S}
}
\IEEEauthorblockA{
Institute for AI Industry Research (AIR), Tsinghua University; Z-Trans AI\\
$^{*}$Co-first authors \quad $^{\dag}$Corresponding authors \quad $^{\S}$ Project lead \quad $^{\ddagger}$Work done during a visit to AIR, Tsinghua\\
\href{https://air-embodied-brain.github.io/Zeva}{Project Page: https://air-embodied-brain.github.io/Zeva} 
}
}

\maketitle
\begin{strip}
    \centering
    \vspace{-15mm}
    \includegraphics[width=\textwidth]{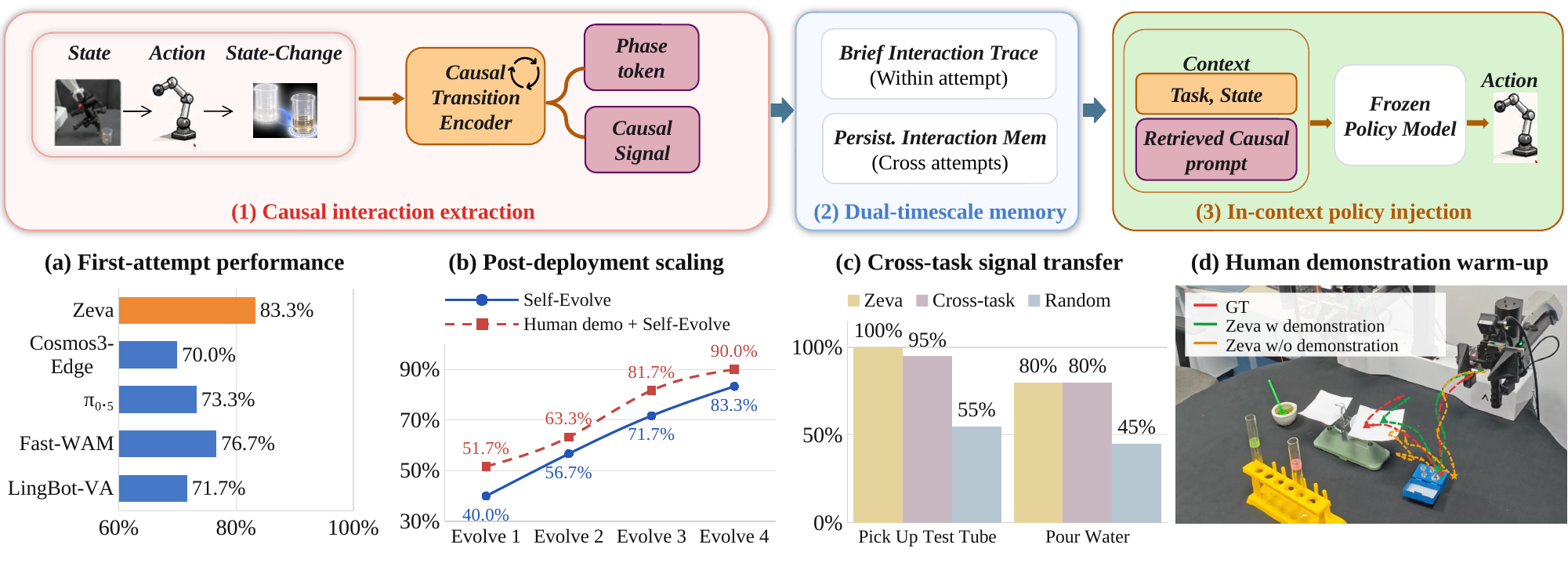}
    \refstepcounter{figure}\label{fig:teaser}
    \par\vspace{1mm}\footnotesize\raggedright
    Fig.~\thefigure.\quad \name enables self-evolving embodied intelligence through In-Context Causal Learning (ICCL).
\name learns the causal relationships between robot actions and resulting state changes by extracting action-induced causal transitions, maintaining dual-timescale interaction memory, and retrieving causal context for in-context policy adaptation. Without parameter updates, \name enables frozen policies to progressively improve through self-evolution and human-guided experience injection, achieving strong real-world performance and cross-task knowledge transfer beyond task-specific demonstrations.\par
\end{strip}

\begin{abstract}
Generalizable embodied manipulation remains difficult to achieve through pretraining alone, due to unseen physical conditions in the real world. We argue that robots need to learn from their own physical interactions on the fly during real-world deployment and use this knowledge to inform subsequent actions. We present \name, the first framework that enables in-context learning from a robot's own physical interaction experience while keeping the policy model frozen. \name employs a Causal Interaction Extractor to encode an executed action and its induced state change into a causal interaction signal, which is stored in a dual-timescale causal memory. For subsequent actions, relevant causal interaction signals are retrieved from memory and injected into the frozen policy model as context. Experiments in simulation and real-world manipulation demonstrate that \name achieves the best performance among the compared frontier VLAs and WAMs and, more importantly, enables self-evolution during deployment without gradient updates. Its success rate continues to improve as the robot accumulates interaction experience. Furthermore, the acquired interaction experience can generalize across tasks.
\end{abstract}


\section{Introduction}

Generalizable interaction-intensive manipulation is the central challenge for embodied AI. 
Embodied foundation models, including VLAs and WAMs, have made rapid progress towards this goal~\cite{openvla,pi0,rt1,rt2,octo,diffusion_policy,cosmos_policy,lingbot_va,dreamzero}, yet deployment in the physical world remains brittle. At deployment, robots inevitably encounter physical conditions that are absent during pretraining, such as novel object geometries, contact dynamics, visual variations, and embodiment-specific execution errors. These conditions interact with continuous, high-dimensional robot states and actions, causing similar actions to produce significant different outcomes. It is therefore difficult for pretraining alone to cover the diversity of physical interactions.

To realize generalizable embodied manipulation, we argue that \textit{the robot should learn from its own physical interaction experiences on the fly during execution, i.e., learning the causality between its action and the induced state change, and can apply it to subsequent actions}. Recent post-pretraining adaptation methods have begun to move in this direction, but still leave gaps. Test-time training methods, such as RoboTTT~\cite{robottt} and WAM-TTT~\cite{wamttt}, convert deployment histories into adaptive fast weights or memory modules through test-time gradient updates. This leaves the action and state-change structure of each interaction implicit. Recent in-context robot learners, such as GEN-1.5~\cite{gen15} and Skild S1~\cite{skild_s1}, instead keep model weights fixed and infer a new task from one or a few demonstrations in context. However, this form of in-context learning primarily targets task generalization rather than on-the-fly interaction learning.


We present \textit{\name}, the first framework that enables in-context learning from the robot's own physical interaction experiences, toward generalized embodied manipulation in real deployment. \name keeps the policy model frozen at deployment, but extracts the causality between actions and state changes during each action step and retrieves this knowledge as context for subsequent action generation. This enables the same frozen policy to adapt to diverse real-world physical conditions across repeated self-attempts and exhibit self-evolution, i.e., the more it attempts, the higher its success rate becomes.

\name realizes this idea as \textbf{In-Context Causal Learning} (ICCL) through three stages. First, \textbf{Causal Interaction Extraction} uses a Causal Transition Encoder (CTE) to integrate visual latents, action encodings, and observed effects into a Causal Interaction State, which is then projected into a task \textbf{Phase Token} and a \textbf{Causal Interaction Signal}. Second, \textbf{Dual-timescale Causal Memory} organizes these signals for deployment-time adaptation: a \textbf{Brief Interaction Trace} (BIT) captures recent within-attempt dynamics, while a \textbf{Persistent Interaction Memory} (PIM) consolidates useful evidence across attempts within the same episode through similarity-based merging. Third, \textbf{In-Context Policy Injection} retrieves phase-matched interaction evidence and constructs a \textbf{Causal Prompt} for the frozen foundation policy. During deployment, all model parameters remain frozen; only the interaction memories are updated online.

We evaluate \name on RoboCasa365-Atomic5 and a real-world chemical manipulation benchmark, \textit{ChemLab-Evo}, using an ARX manipulator. The main results are: (1) \name achieves the best success rate among the compared frontier VLA and WAM models on RoboCasa365-Atomic5, reaching 76.8\%. (2) \name achieves the best success rate on the real-world ChemLab-Evo across all three difficulty levels. (3) \name exhibits \textit{success-rate scaling from its own interaction experience}: on RoboCasa365-Atomic5, the cumulative success rate increases from 26\% at the first attempt to 73\% within four repeated attempts, and ChemLab-Evo shows the same scaling trend across attempts. (4) \name also exhibits in-context learning from human teleoperation demonstrations, further improving performance by up-to 15\% beyond learning from its own interactions alone.

Our contributions are summarized as follows:
\begin{itemize}
    \item We formulate embodied action generalization in real-world deployment as In-Context Causal Learning, where a frozen robot policy learns from its own physical interaction experiences without test-time weight updates.
    \item We propose \name, which combines Causal Interaction Extraction, Dual-timescale Causal Memory, and In-Context Policy Injection for gradient-free self-evolution across repeated attempts.
    \item We validate \name in simulation and real-world manipulation, showing improved success across robot attempts and the strongest success rate compared to frontier embodied policy models.
\end{itemize}

\section{Related Work}
\begin{figure*}[t]
  \centering
  \includegraphics[width=0.9\linewidth]
    {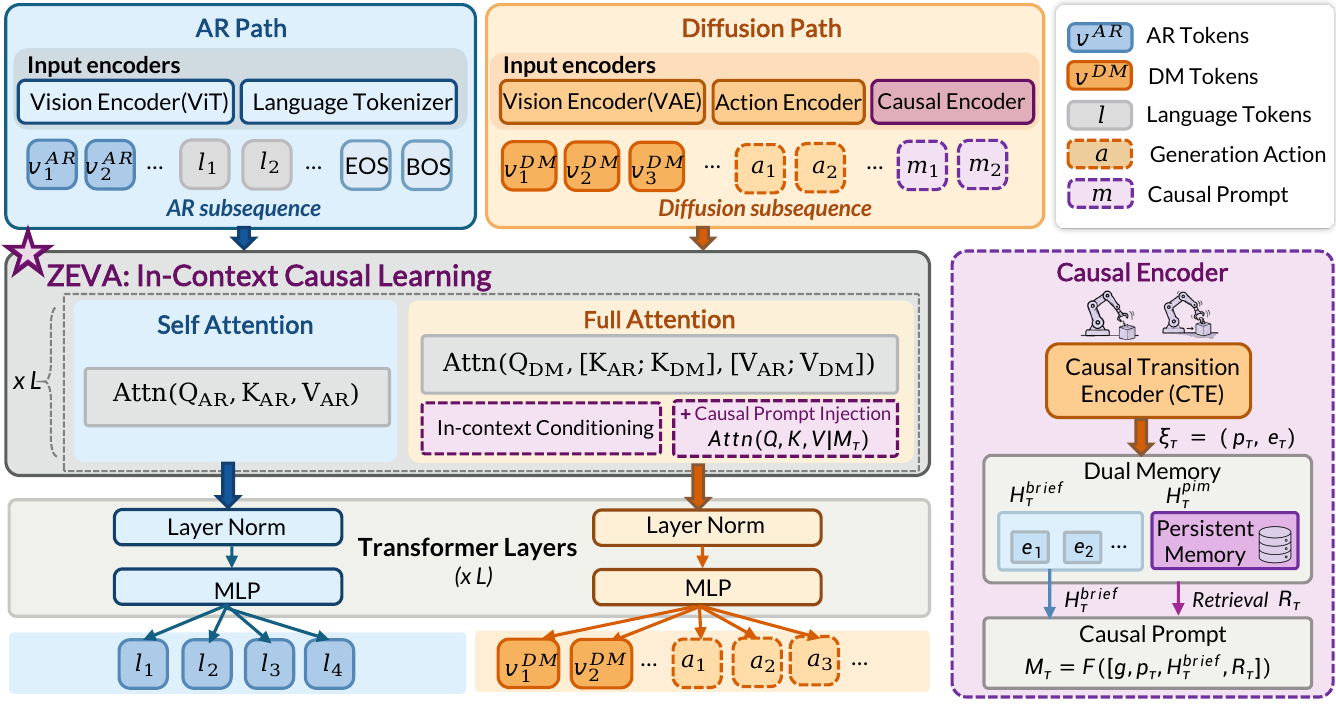}
    \caption{Detailed architecture of the Zeva framework. Zeva integrates causal interaction intelligence into a frozen foundation policy through three main components: (1) Causal Transition Encoder, which maps action-effect pairs into latent causal signals; (2) Dual-timescale Causal Memory, consisting of a Brief Interaction Trace (BIT) for within-attempt dynamics and a Persistent Interaction Memory (PIM) for cross-attempt experience; and (3) Causal Prompt Injection, where retrieved phase-matched evidence is injected into the Transformer’s self-attention layers as an in-context prompt to guide the Diffusion-based action generation.}
    \label{fig:zeva-architecture}
\end{figure*}


\subsection{Vision-Language-Action and World-Action Foundation Models}

VLA models combine a vision-language backbone with an action-generation head, learning an observation-to-action mapping directly from demonstration data; representative work includes OpenVLA~\cite{openvla}, $\pi_0$~\cite{pi0}, and its successors $\pi_{0.5}$~\cite{pi05} and $\pi^{*}_{0.6}$~\cite{recap}. World-Action Models (WAMs) further unify action generation with future visual prediction, absorbing physical priors through large-scale video pretraining~\cite{gr1,gr2}, adapting pretrained video diffusion models to jointly denoise future frames and actions~\cite{cosmos_policy,dreamzero}, or autoregressively interleaving video and action tokens under a causal attention mask~\cite{lingbot_va}. All of these models require large-scale pretraining to obtain generalizable priors, and their weights remain fixed after deployment, so their capability ceiling is set by the pretraining corpus. This work adopts Cosmos3's~\cite{cosmos3} rectified-flow action-generation head as the policy backbone, but obtains continual adaptation after deployment through an external self-evolving mechanism rather than through pretraining scale.



\subsection{Behavioral and Latent-Action Representation Learning}

To mitigate distribution shift caused by high-dimensional observation-to-action mapping, one line of work~\cite{act,vqbet,fast,rth,pact,aem,ding2025adanav,ding2026memcompiler,li2026oxygen,ju2026embodiskill}, learns low-dimensional behavioral representations:  ALAM~\cite{alam} and LAPA~\cite{lapa} learn latent-action encodings from unlabeled video via algebraic-consistency constraints and inverse-dynamics reconstruction, respectively; BehaviorVLA~\cite{behaviorvla} encodes a trajectory into a scene-agnostic global prototype and an online-updated local phase; UniVLA~\cite{univla} derives task-centric latent action representations from action-label-free video across arbitrary embodiments and viewpoints. These methods show that low-dimensional behavioral representations outperform high-dimensional direct mapping, but in every case the representation space is fixed before deployment and cannot accumulate interaction experience from real deployment, making them difficult to keep effective as the distribution keeps changing.

\subsection{Memory-Augmented Robot Policies}

Another line of work equips policies with online memory to handle long-horizon dependencies. MemoryVLA~\cite{memoryvla} maintains a perceptual-cognitive memory bank and handles non-Markovian control through a retrieve-fuse-consolidate mechanism; MEM~\cite{mem} combines short-term video memory with a long-term language-based event summary to support tasks lasting more than ten minutes; DIM-WAM~\cite{dimwam} and MemoryWAM~\cite{memorywam} compress visual history through, respectively, a multi-type historical event bank and a sliding-window-plus-anchor-frame mechanism. In each case, the memory content consists of observation- or semantic-level historical fragments. Determining which direction an action shifted the state on a failed attempt, and how it should be adjusted on the next attempt of the same episode, requires explicitly preserving the action--state-change relationship across attempt boundaries, something perception-side history alone does not capture.

\subsection{Deployment-Time Adaptation and Experience-Driven Self-Improvement}


The line of work closest to our motivation enables policies to improve from their own deployment experience. $\pi^{*}_{0.6}$ (RECAP)~\cite{recap} and VLA-RL~\cite{vla_rl} update policies through offline or online reinforcement learning on autonomous rollouts. However, their gradient-based optimization cycles are too slow to adapt the policy between consecutive attempts and are vulnerable to capability collapse. RoboTTT~\cite{robottt} and WAM-TTT~\cite{wamttt} accelerate adaptation by updating lightweight fast weights or memory modules via test-time gradients. Nevertheless, the adaptation latency remains far beyond the timescale of individual attempts, and WAM-TTT additionally relies on human demonstration videos as the update source, preventing the agent from autonomously exploring and improving through its own interaction experiences.

\section{Method}

\subsection{Problem Formulation: In-Context Causal Learning}

\label{sec:problem-formulation}
We consider an embodied agent~\cite{zetta} executing tasks $\mathcal{T} \sim P(\mathcal{T})$ under varying physical conditions. Our goal is \textbf{In-Context Causal Learning (ICCL)}: enabling the agent to learn from how its actions affect the environment and use this experience to guide subsequent behavior, while keeping model parameters fixed during deployment ($\theta=\theta_0$).

Within an episode, the agent may make multiple attempts at the same task instance. At each decision, the policy uses a causal context $\mathcal{M}_\tau$ built from completed interactions in the current and earlier attempts. This context acts as a causal prompt: adaptation occurs through accumulated interaction evidence rather than deployment-time weight updates.

\textbf{Causal Context as Interaction Evidence}
We represent experience as a sequence of \textbf{interaction triplets}: a visual state $s_\tau$, the action chunk $u_\tau$ executed from that state, and its observed effect $\Delta s_\tau=s_{\tau+1}-s_\tau$. Here $\tau$ indexes action-chunk boundaries. Once the effect is observed, the triplet is encoded into a causal interaction unit:
\begin{equation}
\xi_{\tau+1} = \phi(s_\tau, u_\tau, \Delta s_\tau),
\label{eq:interaction-unit}
\end{equation}
The causal context $\mathcal{M}$ is organized from these units $\xi$. By conditioning on $\mathcal{M}$, the frozen policy $\pi(a_\tau \mid o_\tau, g, \mathcal{M})$ uses observed interaction outcomes to adapt to task-relevant dynamics and constraints.


\subsection{Overview of Zeva}
To realize the ICCL objective, we propose \textbf{Zeva}, a framework that transforms raw interactions into a dual timescale causal memory. Zeva consists of three parts: (1) \textbf{Causal Interaction Extraction}, which maps actions and effects into a latent space, (2) \textbf{Dual timescale Memory Management}, which structures experiences for retrieval, and (3) \textbf{In-Context Policy Injection}, which provides the causal prompt to the foundation model.

Figure~\ref{fig:zeva-architecture} shows how these components couple to the
two-path foundation policy. The autoregressive path encodes multimodal context,
while the diffusion path generates continuous robot actions. For every executed
transition, CTE extracts a phase token and a causal interaction signal: the
Brief Interaction Trace supplies recent within-attempt evidence, whereas the
Persistent Interaction Memory retrieves phase-matched evidence accumulated
across attempts. The resulting causal prompt conditions the full-attention
blocks of the diffusion path, allowing the policy to change its action strategy
through context while keeping all model parameters fixed.

\subsection{Part 1: Causal Interaction Extraction}
To bridge the gap between raw pixels and causal reasoning, we use a \textbf{Causal Transition Encoder}(CTE) to map interactions into a latent causal space. 


\paragraph{Interaction Recurrence}
At each boundary $\tau\ge1$, we use visual latents $s_\tau$, the preceding executed controls $u_{\tau-1}$, and their observed effect $d_\tau=\Delta s_{\tau-1}=s_\tau-s_{\tau-1}$. Inspired by trajectory-level behavioral representations~\cite{behaviorvla},
we integrate these signals into a \textbf{Causal Interaction State} $B_\tau$
via a gated recurrence:
\begin{equation}
B_\tau = \operatorname{GRU}(B_{\tau-1}, [s_\tau;u_{\tau-1};d_\tau]),
\label{eq:original-cte-recurrence}
\end{equation}
with $B_0=F_{\mathrm{init}}(g,s_0)$; brackets denote concatenated input encodings. The state $B_\tau$ is projected into two components: a \textbf{Phase Token} $p_\tau$ representing task progress and a \textbf{Causal Interaction Signal} $e_\tau$ characterizing the dynamics. Together, they form the causal interaction unit $\xi_\tau = (p_\tau, e_\tau)$ as defined in Section~\ref{sec:problem-formulation}.


\paragraph{Learning Objectives} CTE learns interaction evidence through three complementary objectives:
\begin{itemize}
    \item \textbf{Causal Effect Prediction ($\mathcal{L}_{\text{effect}}$)}: We decode pre/post interaction representations to predict visual changes $\Phi(o_{\tau+1})-\Phi(o_\tau)$ from a fixed encoder, using mean-squared error $\mathcal{L}_{\mathrm{MSE}}$. A shared prediction head maps an action-conditioned pre-interaction code from $(B_\tau,u_\tau)$ and the post-interaction signal $e_{\tau+1}$ to visual-change estimates:
    \begin{equation}
    \mathcal{L}_{\mathrm{effect}}=
    \alpha_m\mathcal{L}_{\mathrm{MSE}}
    +\mathcal{L}_{\mathrm{aux}}.
    \label{eq:effect-total}
    \end{equation}
    The weighted auxiliary term $\mathcal{L}_{\mathrm{aux}}$ combines action-summary reconstruction, pre/post alignment, and anti-collapse regularization.
    \item \textbf{Task Identity Clustering ($\mathcal{L}_{\text{task}}$)}: Let $z_i$ be the normalized trajectory mean of unit-normalized task projections of $B_\tau$. With same-task positives $\mathcal{P}_i$, supervised contrastive learning gives
    \begin{equation}
    \mathcal{L}_{\mathrm{task}}=-\mathbb{E}_{i,j\in\mathcal{P}_i}
    \log\frac{\exp(z_i^\top z_j/T_t)}
    {\sum_{k\ne i}\exp(z_i^\top z_k/T_t)}.
    \label{eq:task-contrastive}
    \end{equation}
    We average uniformly over anchors with positives, then over their positives, excluding self-pairs. The denominator includes all other trajectories in the minibatch. A separate task projection preserves the distinction between task identity and stored effect signals.
    \item \textbf{Phase Progression ($\mathcal{L}_{\text{phase}}$)}: Let $\bar p_{i,t}$ be a normalized phase token and $v_i$ the unit direction from the first to the last token of trajectory $i$. We encourage ordered progress along this direction:
    \begin{equation}
    \mathcal{L}_{\mathrm{phase}}=\mathbb{E}_{i,t}
    \bigl[m-v_i^\top(\bar p_{i,t+1}-\bar p_{i,t})\bigr]_+,
    \label{eq:phase-progression}
    \end{equation}
    where $[x]_+=\max(0,x)$ and $m>0$. The average is over valid trajectories, then adjacent token pairs within each trajectory.
\end{itemize}
The combined representation objective is
\begin{equation}
\mathcal{L}_{\mathrm{CTE}}=
\lambda_e\mathcal{L}_{\mathrm{effect}}+
\lambda_t\mathcal{L}_{\mathrm{task}}+
\lambda_p\mathcal{L}_{\mathrm{phase}},
\label{eq:original-cte-total}
\end{equation}
Weights are nonnegative and $T_t>0$. Auxiliary heads are used only for training; deployment retains the recurrent CTE and its phase/signal projections.

\begin{algorithm}[t]
\caption{In-Context Causal Inference with Zeva}
\label{alg:cosmos_evo_inference}
\footnotesize
\begin{algorithmic}[1]

\Require Task instruction $g$, environment $\mathcal{E}$, maximum attempts $R\ge1$, horizon $T$, retrieval size $K$, brief-memory window $K_b$, stop-on-success flag $b$
\Require Frozen Causal Transition Encoder (CTE), Memory Context Encoder $F_{\text{mem}}$, and foundation policy $\pi$
\Ensure Task outcome and updated persistent interaction memory $\mathcal{H}^{\mathrm{pim}}$

\State \textbf{Freeze} all neural parameters $\theta$
\State $\mathcal{H}^{\mathrm{pim}} \gets \varnothing$ \Comment{Episode-local; optional demonstration initialization}

\For{$r=1,\ldots,R$}

    \State $o_0 \gets \Call{Reset}{\mathcal{E}}$; encode visual state $s_0$
    \State $\mathcal{H}^{\mathrm{brief}} \gets \varnothing$ \Comment{Local context reset per attempt}
    \State $B_0 \gets F_{\text{init}}(g, s_0)$ \Comment{Initialize causal interaction state}
    \State $p_0 \gets P_{\text{phase}}(B_0)$, $e_0\gets P_{\text{signal}}(B_0)$

    \For{$t=0,\ldots,T-1$}

        \State $\mathcal{R}_t \gets \Call{PhaseRetrieval}{p_t, \mathcal{H}^{\mathrm{pim}}, K}$ \Comment{Retrieve relevant interaction evidence}

        \State Construct $\mathcal{M}_t$ from $g,p_t,\mathcal{H}^{\mathrm{brief}},\mathcal{R}_t$ using the shared history projector and $F_{\mathrm{mem}}$

        \State $a_t \sim \pi(\cdot \mid o_t, g, \mathcal{M}_t)$ \Comment{Implicit causal inference and action generation}

        \State $(o_{t+1},u_t,\texttt{terminal})\gets\Call{Step}{\mathcal{E},a_t}$
        \State Encode $s_{t+1}$; $d_{t+1}\gets s_{t+1}-s_t$ \Comment{$u_t$: executed controls}

        \State $B_{t+1}\gets\operatorname{GRU}(B_t,[s_{t+1};u_t;d_{t+1}])$
        \State $p_{t+1} \gets P_{\text{phase}}(B_{t+1})$, $e_{t+1} \gets P_{\text{signal}}(B_{t+1})$

        \State $\mathcal{H}^{\mathrm{brief}} \gets \Call{UpdateBrief}{\mathcal{H}^{\mathrm{brief}}, e_{t+1}, K_b}$ \Comment{Update sliding window}

        \State $\mathcal{H}^{\mathrm{pim}} \gets \Call{CausalMerge}{\mathcal{H}^{\mathrm{pim}}, p_{t+1}, e_{t+1}}$

        \If{$\texttt{terminal}$}
            \State \textbf{break}
        \EndIf

    \EndFor

    \State $y_r\gets\Call{Outcome}{\mathcal{E}}$ \Comment{Success or failure of this attempt}
    \If{$b=1$ \textbf{and} $y_r=\textsc{Success}$}
        \State \Return $(y_r, r, \mathcal{H}^{\mathrm{pim}})$
    \EndIf

\EndFor

\State \Return $(y_R, R, \mathcal{H}^{\mathrm{pim}})$

\end{algorithmic}
\end{algorithm}

\subsection{Part 2: Dual timescale Causal Memory}
Zeva organizes the phase--effect units in~\eqref{eq:interaction-unit} into two memory streams to support adaptation within and across attempts of the current episode.

\paragraph{Brief Interaction Trace} $\mathcal{H}_\tau^{\text{brief}}$ stores at most $K_b$ completed interaction signals in temporal order, providing immediate local context:
\begin{equation}
\mathcal{H}_\tau^{\text{brief}} = (e_j)_{j=\max(1,\tau-K_b+1)}^{\tau}.
\end{equation}
The trace starts empty and resets between attempts; the initialization token $e_0$ is not a completed interaction.

\paragraph{Persistent Interaction Memory.} $\mathcal{H}_\tau^{\text{pim}}$ accumulates causal units $\xi = (p, e)$ across attempts of the same episode and is cleared before a new episode begins. To ensure scalability, we employ similarity based merging. For a newly completed unit $\xi_{\tau+1} = (p_{\tau+1}, e_{\tau+1})$, we find the most similar entry $\xi^* \in \mathcal{H}_\tau^{\text{pim}}$ by calculating:
\begin{equation}
S_{\text{merge}}(\xi_{\tau+1}, \xi_j) = \beta_p \text{sim}(p_{\tau+1}, p_j) + \beta_e \text{sim}(e_{\tau+1}, e_j).
\end{equation}
With best-match score $S^*=S_{\text{merge}}(\xi_{\tau+1},\xi^*)$, the persistent memory is updated as follows:
\begin{equation}
\mathcal{H}_{\tau+1}^{\text{pim}} \leftarrow
\begin{cases}
\begin{aligned}
&(\mathcal{H}_{\tau}^{\text{pim}} \setminus \{ \xi^* \}) \\
&\quad \cup \{ \text{Merge}(\xi^*, \xi_{\tau+1}) \}
\end{aligned}
& \text{if } S^* > \tau_{\text{merge}}, \\
\mathcal{H}_{\tau}^{\text{pim}} \cup \{ \xi_{\tau+1} \}, & \text{otherwise}.
\end{cases}
\end{equation}
Here $\text{sim}$ is cosine similarity, $\beta_p,\beta_e\ge0$ sum to one, and $\text{Merge}$ takes count-weighted means of the phase and effect vectors and normalizes each mean. An empty memory directly stores the new unit. This ensures that a new experience is only merged if it matches both the task progress and the physical dynamics of an existing record.

\subsection{Part 3: In-Context Policy Injection}
The final stage transforms stored experiences into a structured causal prompt for the foundation model.

\paragraph{Phase-Conditioned Retrieval} We use the current phase $p_\tau$ to query the persistent memory. The retrieved set $\mathcal{R}_\tau$ consists of signals with the most similar stored phase representations; this is a learned retrieval heuristic rather than an explicit alignment of trajectory stages:
\begin{equation}
\mathcal{R}_\tau=(e_j)_{j\in\mathcal{I}_\tau},\qquad
\mathcal{I}_\tau=\operatorname{TopK}_j\operatorname{sim}(p_\tau,p_j).
\end{equation}
The indices refer to stored records, ordered by similarity. Retrieve all records if fewer than $K$ exist; an empty memory returns an empty sequence.
\paragraph{Causal Prompt Construction} The memory context $\mathcal{M}_\tau$ integrates the task token $g$, phase token $p_\tau$, and both history streams.
\par
Write $\widetilde{\mathcal{H}}_\tau=P_{\mathrm{proj}}(\mathcal{H}_\tau^{\mathrm{brief}})$ and $\widetilde{\mathcal{R}}_\tau=P_{\mathrm{proj}}(\mathcal{R}_\tau)$ for the shared history projections. The Memory Context Encoder fuses them as
\begin{equation}
\mathcal{M}_\tau=F_{\mathrm{mem}}\bigl([g;p_\tau;
\widetilde{\mathcal{H}}_\tau;\widetilde{\mathcal{R}}_\tau]\bigr).
\end{equation}
Here $g$ denotes the embedded instruction; $g$ and $p_\tau$ anchor task and phase, while the history projections provide observed interaction evidence. In Fig.~\ref{fig:zeva-architecture}, the causal encoder maps $\mathcal{M}_\tau$ into prompt tokens for the diffusion path. This \textbf{Causal Prompt} supports implicit reasoning about action outcomes and conditions $\pi(a_\tau\mid o_\tau,g,\mathcal{M}_\tau)$ without deployment-time parameter updates.

\paragraph{Offline Policy Conditioning} CTE is first optimized with~\eqref{eq:original-cte-total} and frozen. The history projector, memory context encoder, prompt mapping, and policy are then optimized with conditional action flow matching. Let $a_\tau^*$ be a successful or corrected action chunk, $\epsilon\sim\mathcal{N}(0,I)$, and $\sigma\sim\mathcal{U}[0,1]$. With context $\mathcal{C}_\tau=(o_\tau,g,\mathcal{M}_\tau)$, define
\begin{equation}
\begin{aligned}
x_\sigma&=(1-\sigma)\epsilon+\sigma a_\tau^*,\\
\mathcal{L}_{\mathrm{policy}}&=
\mathbb{E}_{\tau,\sigma,\epsilon}
\operatorname{MSE}\!\left(v_\vartheta(x_\sigma,\sigma;\mathcal{C}_\tau),
a_\tau^*-\epsilon\right),
\end{aligned}
\label{eq:policy-flow}
\end{equation}
where $v_\vartheta$ is the policy's action velocity field and the error is averaged over valid action coordinates.

\subsection{Inference}
During deployment, the agent reads memory before acting and writes the newly observed effect afterward. Each retry restores the environment's initial state and resets the recurrent state and BIT, retaining only PIM. Algorithm~\ref{alg:cosmos_evo_inference} uses $t$ for the chunk-boundary index $\tau$.

\section{Experiments}
We evaluate Zeva on the RoboCasa365-Atomic5 simulation benchmark and the real-world ChemLab-Evo benchmark. Our experiments are designed to answer three key questions: (1) \textbf{Benchmark Performance}: Can Zeva achieve competitive
task success and long-horizon progress across simulated and real-world manipulation tasks of increasing complexity? (2) \textbf{Post-Deployment In-Context Scaling}: Can accumulated interaction experience and a one-shot human demonstration improve a frozen policy across repeated attempts without parameter updates? (3) \textbf{Causal Memory and Cross-Task Generalization}: Do CTE, BIT, and PIM encode useful action-induced interaction signals that contribute to performance improvement and support meaningful cross-task retrieval and transfer?

\RealWorldResultsTable
\RoboCasaResultsTable

\subsection{Experimental Setup}

\subsubsection{Episode and Attempt Protocol}
We use two evaluation units: \textbf{episode} and \textbf{attempt}. An episode
corresponds to a single task instance defined by its initialization, including
the scene layout, object poses, and the robot's starting configuration. A
\textbf{randomized episode} independently samples this initialization at the
beginning of the episode. A \textbf{fixed episode} selects the initialization
once and holds it unchanged across repeated attempts. An attempt is one
continuous policy execution from the initial observation until termination or
success. Before each new attempt in a fixed episode, the environment is restored
to the same initial state.

The Brief Interaction Trace (BIT) is cleared at the beginning of every
attempt, whereas the Persistent Interaction Memory (PIM) is retained across
attempts within the same episode. Both memories are cleared before the next
episode. An episode terminates immediately after the first
successful attempt, and every compared method receives the same maximum retry
budget.

\subsubsection{Simulation Benchmark}

\textbf{RoboCasa365-Atomic5} (\textbf{Atomic5}) comprises five kitchen tasks: Turn On Electric Kettle, Close Toaster Oven Door, Turn On Microwave, Coffee Setup Mug, and Open Stand Mixer Head. Each task uses 50 randomized episodes evaluated twice; Table~\ref{tab:exp_robocasa} reports the mean SR of the two runs. Zeva uses Cosmos3-Nano~\cite{cosmos3} in simulation.

\subsubsection{Real-World Chemical Lab (ChemLab-Evo)}
\textbf{ChemLab-Evo} uses an \textbf{ARX} manipulator, a 14-DoF bimanual platform equipped with both wrist-mounted and first-person cameras (680 × 340 resolution) in an $80\,\mathrm{cm}\times60\,\mathrm{cm}$ workspace. Its seven tasks span three levels of increasing compositional complexity. Level 1 (Atomic) evaluates Pick Up Test Tube, Place Beaker, and Pour Water. Level 2 (Short-sequence) composes skills through Titration and Prepare Salt Solution. Level 3 (Complex) evaluates long-horizon procedures through Balance Weighing and Extraction. As complexity increases, success shifts from one manipulation primitive to completing all stages in the prescribed order. The main benchmark and each ablation configuration use 20 independently randomized episodes per evaluated task. To enable deployment on edge devices (RTX 4090), Zeva uses Cosmos3-Edge~\cite{cosmos3} in real experiments.

\subsubsection{Comparative Baselines}
On Atomic5, we compare Zeva with LingBot-VA~\cite{lingbot_va}, Xiaomi-Robotics-1~\cite{xiaomi_robotics_1}, $\tau_0$-WM~\cite{tau0_wm}, Fast-WAM~\cite{fast_wam}, and Cosmos3-Nano~\cite{cosmos3}. For ChemLab-Evo, we additionally include $\pi_{0.5}$~\cite{pi05}. All baselines are reproduced locally using the same task-specific training datasets, observation and action interfaces, and evaluation protocol as Zeva.

\subsection{Evaluation Metrics}

\textbf{Benchmark Metrics.}
For both simulation and ChemLab-Evo, we report the task-level \textbf{Success Rate (SR)} and the macro-average SR across the evaluated tasks. Each SR is computed over 50 randomized episodes per simulation task and 20 randomized episodes per real-world task. Because binary success provides limited resolution for long-horizon tasks, we additionally report a normalized \textbf{process Score}. For a task with $M_t$ ordered stages, let $L_i$ be the length of the longest stage prefix completed in the prescribed order during episode $i$. The task-level Score is
\[
    \mathrm{Score}_t = \frac{100}{N}\sum_{i=1}^{N}\frac{L_i}{M_t},
\]
where $N=20$ for ChemLab-Evo. An episode receives 100 only if all stages are completed, while a failed episode retains credit for the valid progress made before termination.

\textbf{Post-Deployment Scaling Metrics.}
To quantify how capability scales with accumulated interaction experience, let $y_{i,k} \in \{0,1\}$ denote the outcome of fixed episode $i$ at attempt $k$. We report the \textbf{Cumulative Success Rate (CSR@$K$)}, the fraction of fixed episodes completed within an attempt budget $K$:
\[
    CSR@K = \frac{1}{N}\sum_{i=1}^{N}
    \mathbb{I}\!\left[\max_{1\leq k\leq K} y_{i,k}=1\right].
\]

\subsection{Experiment Results}

\subsubsection{Real-World Evaluation on ChemLab-Evo}

We further evaluate Zeva on the real-world ChemLab-Evo benchmark using an ARX manipulator. The SR block of Table~\ref{tab:exp_real} is organized by task complexity, from atomic manipulation primitives to multi-stage chemical procedures. Zeva achieves the best average success rate at every complexity level. Relative to the strongest baseline in each level, it improves the atomic, short-sequence, and complex averages by 6.6, 5.0, and 5.0 percentage points, respectively. Since terminal success alone obscures partial progress on the two long-horizon tasks, the process-score block on the right further compares their ordered stage completion. Zeva obtains Scores of 47.50 on Balance Weighing and 67.14 on Extraction, exceeding the strongest baseline for each task by 8.75 and 8.57 points, respectively. Its average Score of 57.32 improves over the best baseline average by 12.59 points.

\subsubsection{Simulation Benchmark Results}

As shown in Table~\ref{tab:exp_robocasa}, Zeva achieves the highest average success rate of 76.8\%, exceeding Fast-WAM by 4.4 percentage points.

\subsection{Ablation Studies}
We ablate the two interaction-memory timescales on five real-world
ChemLab-Evo tasks: all three atomic tasks and both short-sequence tasks.
Table~\ref{tab:ablation} compares the full model with variants that remove
the Brief Interaction Trace, the Persistent Interaction Memory, or both.
We denote these components as BIT and PIM, respectively, and report the
task-level SR over 20 randomized episodes for each configuration.

The full model performs best on all five tasks. Removing PIM reduces the
SR by 10--20 percentage points, while removing BIT causes a larger decrease of
15--30 percentage points, with the largest drops on Pour Water and Titration.
These controlled results support the complementary roles
of cross-attempt memory and within-attempt context. Disabling both components
yields the weakest result on every task.

\AblationResultsTable
\vspace{1.5mm}

\raggedbottom
\subsection{Post-Deployment In-Context Scaling}
\subsubsection{Post-Deployment Capability Scaling}
\begin{figure}[h]
  \centering
  \includegraphics[width=0.75\linewidth]
    {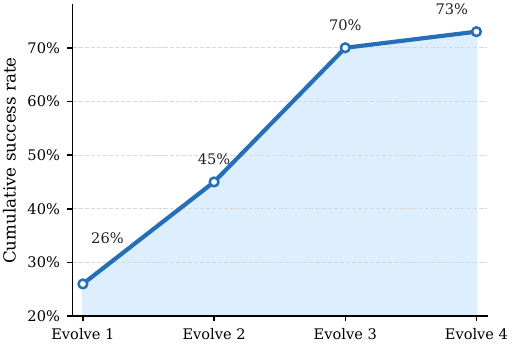}
  \caption{Post-deployment capability scaling on RoboCasa365.
  Each of the five tasks contributes 20 randomized episodes,
  yielding 100 equally weighted task--seed pairs. Evolve 1–4 denote selected evolution milestones.}
  \label{fig:five-task-pooled-csr}
\end{figure}

\begin{figure}[h]
  \centering
  \includegraphics[width=0.88\linewidth]
    {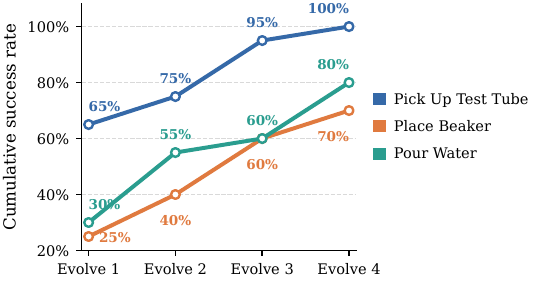}
  \caption{Post-deployment capability scaling on three ChemLab-Evo tasks using separately selected fixed episodes. Evolve 1–4 denote selected evolution milestones.}
  \label{fig:real-world-scaling}
\end{figure}

We next examine whether Zeva improves its probability of task completion as interaction experience accumulates across repeated attempts. This analysis is conducted on fixed episodes selected separately from the randomized episodes used in the main benchmark tables. For Atomic5, the scaling subset contains 20 fixed episodes per task, yielding 100 equally weighted task--configuration pairs; it is distinct from the standard evaluation above, which uses 50 randomized episodes per task across a broader set of initial configurations. Figure~\ref{fig:five-task-pooled-csr} reports four milestones along the evolution trajectory, denoted Evolve 1--4. These labels index selected stages of evolution rather than consecutive attempt numbers. The pooled cumulative success rate rises from 26\% at Evolve 1 to 45\% at Evolve 2, reaches 70\% at Evolve 3, and plateaus at 73\% by Evolve 4.
On ChemLab-Evo, Figure~\ref{fig:real-world-scaling} likewise uses separately selected fixed episodes and reports four evolution milestones for the three atomic tasks. The plotted fixed episodes are independent of the randomized episodes in Table~\ref{tab:exp_real}. Across these milestones, cumulative success increases monotonically from 65\% to 100\% for Pick Up Test Tube, from 25\% to 70\% for Place Beaker, and from 30\% to 80\% for Pour Water.

\subsubsection{One-Shot Warm-Up from a Human Demonstration}
We further present a one-shot memory warm-up case study on the Balance Weighing task. A human physically guides the ARX manipulator once to pick a calibration weight and place it on the balance; CTE encodes the resulting action--state-change trajectory to initialize PIM while the policy remains frozen. We compare the resulting Zeva execution with a baseline that uses the same frozen policy without memory warm-up. As shown in Figure~\ref{fig:human-warmup}, Zeva reproduces the demonstrated approach, grasp, transfer, and placement stages. Its end-effector trajectory reaches both the grasp and placement events, whereas the baseline deviates after approaching the calibration weights and does not complete a valid placement. This case illustrates that a single human experience can warm-start Zeva through its memory interface without gradient updates before autonomous interaction continues.

\HumanWarmupFigure
\HumanWarmupScalingFigure

We further quantify this warm-up effect on the three atomic ChemLab-Evo tasks.
For every evaluated fixed episode, the warm-up condition provides one
task-matched human demonstration before autonomous self-evolution, while the
comparison condition self-evolves without this initialization. As shown in
Figure~\ref{fig:human-warmup-scaling}, human warm-up improves or matches SR at
every evolution milestone. The gain reaches 20 percentage points on Place
Beaker and 15 points on Pour Water; on Pick Up Test Tube, warm-up improves the
early milestones before both conditions converge to 100\%.

\subsection{Cross-Task Generalization of Causal Interaction Signals}
\subsubsection{Cross-Task Transfer}
We first test whether task-local causal interaction signals can be replaced by
functionally similar signals retrieved from other tasks during policy execution.
For each retrieval, we substitute the corresponding task-local interaction
signal with its nearest cross-task match while keeping the policy frozen and
all remaining inputs unchanged. As shown in
Figure~\ref{fig:cross-task-effect-transfer}, SR changes from 100\% to 95\% on
Pick Up Test Tube and remains 80\% on Pour Water. In contrast, replacing the
task-local signal with a randomly selected cross-task interaction signal
reduces SR to 55\% and 45\%, respectively. Nearest cross-task retrieval
therefore exceeds random replacement by 40 and 35 percentage points while
preserving most or all of the original task performance in these two settings.

\CrossTaskEffectTransferFigure

\CrossTaskEffectSection

\subsubsection{Cross-Task Retrieval}
We next probe whether the CTE causal interaction signal $e_\tau$ captures
action-induced state changes beyond task-specific appearance. For each query
transition, we retrieve nearest neighbors by cosine similarity between
interaction signals from a diagnostic memory pool that contains only other
tasks; this analysis does not alter the task-local memory protocol used for the
quantitative evaluation. Figure~\ref{fig:cross-task-effect-retrieval} shows
representative matches for three effects.

A pouring transition from Pour Water retrieves container-tilting transitions
from Prepare Salt Solution and Extraction. Gripper closure around a test tube
retrieves contact-establishment transitions involving different vessels, while
lifting retrieves post-grasp upward motions across object categories. Thus,
transitions with different objects, viewpoints, and task instructions are
aligned by functionally equivalent physical effects, providing a mechanism for
future cross-task memory reuse.

\subsection{Qualitative and Representation Analysis}
\subsubsection{Case-Wise Post-Deployment Evolution}
Figure~\ref{fig:qualitative-executions} complements the aggregate scaling curves
with case-wise evidence of how interaction memory changes later retries and the
eventual successful execution. The four frames in each row are arranged
chronologically, but they do not represent four distinct attempts.

\CaseEvolutionFigure

\noindent\textbf{Pick Up Test Tube.}
In the first two attempts, the gripper
reaches the rack but does not establish a stable grasp. In the subsequent
successful attempt, the third frame shows the corrected grasp, while the fourth
shows the robot retracting with the tube during the same continuous execution.
This progression is consistent with the cross-attempt use of the action--effect
evidence retained in PIM.

\noindent\textbf{Place Beaker.}
The first two attempts expose errors in wrist orientation
and placement near the target region. The final two frames then show two stages
of one successful attempt: Zeva first secures the beaker and subsequently places
it at the target. This corrected execution illustrates the type of cross-attempt
adaptation supported by the retained interaction evidence.

\noindent\textbf{Pour Water.}
This task additionally requires coordinated
transport, rim alignment, and controlled rotation. Across the first three
failed attempts, the observed behavior reveals remaining relative-pose and tilt
errors. In the subsequent successful attempt, Zeva positions the tube over the
beaker and executes the required pouring motion. This qualitative sequence is
consistent with progressive correction from accumulated interaction evidence.

\subsubsection{CTE Representation Analysis}
We further use t-SNE to visualize the interaction embeddings learned by CTE
across all seven ChemLab-Evo tasks. Figure~\ref{fig:cte-tsne} compares the full
three-stream representation with a variant that removes the effect stream.
Full CTE produces compact, task-specific clusters with clear separation,
whereas removing the effect stream increases within-task dispersion and
inter-task mixing. This comparison is consistent with observed state changes
contributing to task-discriminative causal interaction memory.

\CTERepresentationFigure

\flushbottom
\section{Conclusion}
\name is the first framework to enable in-context learning of interaction causality from a robot's own physical experience for generalizable embodied manipulation, while keeping the policy model frozen. Experiments in simulation and real-world manipulation demonstrate that \name achieves the best performance among the frontier VLA and WAM models. Moreover, its success rate consistently improves over repeated attempts as interaction experience accumulates. 

A current limitation is that \name learns causality only from attempts generated for task execution, rather than actively collecting interactions for learning. Future work will therefore investigate causality-driven exploration, in which the robot purposefully selects informative rollouts to reduce uncertainty about physical interactions and acquire experience more efficiently.

\section*{Acknowledgment}
We thank An Pan, Zexu Wang, Yi Tao, Zijian Wang, Shuhao Wu, Wenhui Gu, and Bowen Yang for their contributions on engineering and demos.

\bibliographystyle{IEEEtran}
\bibliography{refs}

\end{document}